\documentclass{article}

\usepackage[final]{corl_2026}

\usepackage{amsmath}
\usepackage{graphicx}
\usepackage{booktabs}
\usepackage{xltabular}
\usepackage{array}
\usepackage{pdflscape}
\usepackage{tikz}
\usetikzlibrary{positioning}
\usepackage{placeins}
\usepackage{wrapfig}
\usepackage{makecell}
\usepackage{subcaption}
\usepackage{float}
\usepackage{multirow}

\title{COOL: Curiosity-Driven Object Ownership Learning for Personalized Robotic Assistance}

\author{%
  \textbf{Samira Huber}\thanks{Equal contribution} \qquad 
  \textbf{Ruben Hammele}\footnotemark[1] \qquad 
  \textbf{S\"oren Pirk} \\
  Department of Computer Science, Kiel University, Germany \\
  \texttt{\{samira.huber, ruben.hammele\}@cs.uni-kiel.de}, \texttt{soeren.pirk@gmail.com}
}

\newcommand{\papername}{COOL}

\begin{document}

\maketitle

\begin{figure}[h]
    \centering
    \includegraphics[width=1\textwidth]{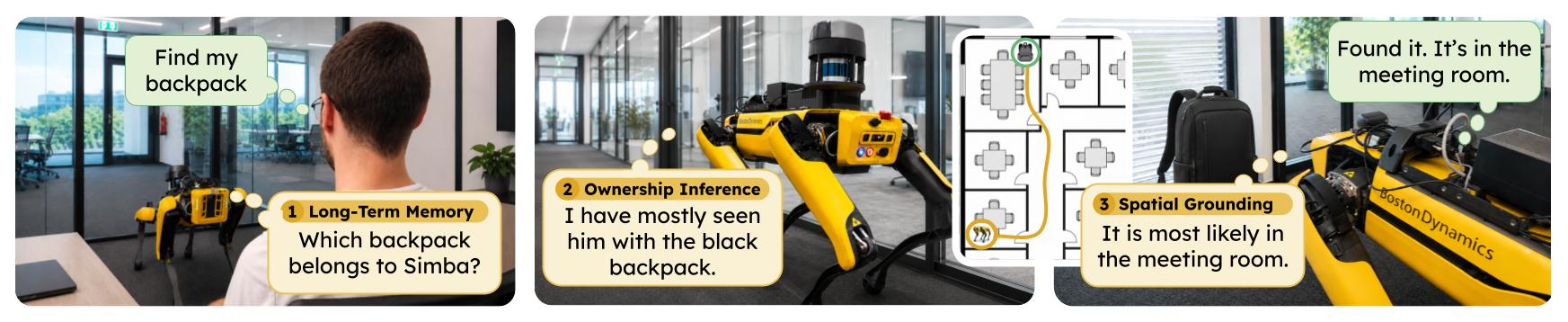}
    \vspace{-4mm}
    \caption{\papername{} enables personalized object search by combining long-term memory, ownership inference, and spatial grounding. Given a natural-language request such as “Find my backpack”~(left), the robot first retrieves prior observations linking people and object instances, then infers the likely owner from accumulated human--object interaction evidence~(middle), and finally grounds the selected object in the map to navigate to its most likely location~(right).}
    \label{fig:cover-image}
\end{figure}

\begin{abstract}
Robots are increasingly expected to provide personalized services in everyday environments. To do so, they must ground natural-language commands such as \textit{Where is my backpack?} or \textit{Find my bottle} and execute them by reasoning about object instances, people, locations, and ownership. This is challenging because ownership is rarely labeled explicitly and must be inferred from long-term, behavioral evidence of human--object interactions. To address this, we present \textit{\papername}, a novel robotic framework for autonomously learning object ownership from everyday observations and maintaining a long-term spatial memory of its environment. To keep its memory current, \textit{\papername} uses an agent-based curiosity-driven data collection strategy that guides the robot toward the most promising locations to gain information and refresh stale observations. Offline experiments, ablation studies, and real-world evaluations show that \textit{\papername} can infer ownership relations from real-world interactions and use this knowledge for ownership-conditioned navigation and task execution. Code, prompts, and evaluation data are available at: \url{https://samirahuber.github.io/cool}.
\end{abstract}


\section{Introduction}
\label{sec:introduction}

In everyday human environments, robots are increasingly tasked with providing personalized services through natural-language commands. In such settings, users may issue instructions such as \textit{Where is Simba's backpack?} or \textit{Find my bottle}. Executing such commands requires more than object detection and navigation: the robot must identify the intended object instance, determine who it is associated with, and know where it is likely to be located. 
This is particularly challenging in shared environments such as offices or homes, where ownership is often ambiguous because objects may be visually similar, shared, borrowed, or moved over time.
Ownership is rarely directly observable: people do not continuously label their belongings, and ownership can become ambiguous when objects are shared, borrowed, or moved. Object ownership must therefore be inferred indirectly from repeated observations, spatial context, and human--object interactions. A service robot must collect and maintain long-term evidence about people, objects, places, and interactions.

Existing work addresses parts of this problem, but not the full ownership-learning pipeline. 
Memory-based navigation systems store spatial, semantic, or temporal observations for object search and embodied question answering, but do not model persistent ownership relations~\citep{niu2026ssmgnav,anwar2025remembr,huber2026humembr,ginting2025mindpalace,saxena2025grapheqa,zhang2026fasteqa,monaci2025rana}. Conversely, personalized object-search systems often require explicit user input, demonstrations, or visual references~\citep{IJAC-2019-07-140,ahfe2023interactiveOwnership,dai2023think,findthis2023,barsellotti2024pin,khanna2024goatbench,kim2024pga,paramonov2024swissdino}. What remains missing is a robotic system that autonomously learns ownership from long-term observations and uses it for task execution.

In this paper, we address this gap by treating ownership as a latent relation inferred from long-term robot observations rather than as an explicit label. Our framework associates repeated detections of people and objects, records human--object interactions, and converts accumulated evidence into ownership-conditioned actions for personalized object search.
Specifically, we present \textit{\papername} (\textit{\underline{C}uriosity-Driven \underline{O}bject \underline{O}wnership \underline{L}earning for Personalized Robotic Assistance}), an integrated framework for ownership-aware object search in shared human environments. \textit{\papername} maintains a long-term spatial memory that links people, object instances, locations, and interaction histories over time. It continuously clusters repeated observations using re-identification, detects human--object interactions as ownership evidence, and records where people and objects have been observed. A curiosity-driven agent guides the robot toward under-observed areas to refresh stale or incomplete memory. At query time, a user-facing agent resolves commands such as \textit{Find Simba’s backpack} by retrieving relevant observations, inferring likely person–object relations, and selecting the corresponding object instance (see Fig.~\ref{fig:cover-image}).

We evaluate \textit{\papername} through offline experiments, ablations, and real-world tests in an office environment. These experiments assess the framework's ability to associate repeated observations of the same people and objects and infer ownership relations from human--object interaction evidence. It updates object locations over time and uses this knowledge for object search and navigation. Combining long-term spatial memory, re-identification, interaction detection, and active data collection enables robots to autonomously acquire ownership knowledge and apply it to personalized tasks.

In summary, the main contributions of this paper are:
(1) \textbf{Ownership inference}: a dynamic framework for inferring person--object ownership from long-term robot observations, without explicit labels, demonstrations, or dialogue.
(2) \textbf{Persistent spatial memory}: a structured memory linking people, object instances, locations, and human--object interaction histories.
(3) \textbf{Ownership-conditioned task execution}: a tool-using agent that answers ownership-conditioned questions and executes navigation commands from memory.
(4) \textbf{Memory refresh}: an autonomous data-collection agent that seeks informative locations and refreshes stale observations.

\section{Related Work}
\label{sec:realted_work}

\textbf{Memory-based Navigation.}
Recent work gives embodied agents long-term memory for object search, exploration, and embodied question answering. 
These methods store prior observations as maps, graphs, vector memories, or scene graphs, and often combine them with large language models (LLMs) or vision-language models (VLMs) for retrieval and reasoning~\citep{niu2026ssmgnav,anwar2025remembr,huber2026humembr,ginting2025mindpalace,saxena2025grapheqa,zhang2026fasteqa}. 
Retrieval-augmented navigation and multimodal retrieval-augmented generation (RAG) reuse prior episodes or perceptual streams for navigation and temporal reasoning, while general LLM memory systems focus on textual recall and memory consolidation~\citep{monaci2025rana,luo2025videorag,pan2025secom,du2026memguide,xu2025amem,chhikara2025mem0}. Unlike prior work, \papername{} links persistent people, objects, locations, and interaction histories to infer person--object ownership relations for personalized robot assistance.\\
\textbf{Re-Identification.}
Visual re-identification links observations of the same entity across time using image embeddings. Many general-purpose backbones provide strong visual features~\citep{convnext,vit,efficientnetv2,dinov3}, while 
Person ReID models such as OSNet are optimized for identity matching~\citep{torchreid,zhou2019osnet,zhou2021osnet}. 
Tracking-by-detection methods link detections across nearby frames using motion, appearance, and confidence cues~\citep{zhang2022bytetrack,strongsort,efstrongsort}. 
Recent work improves robustness under deployment shift, heterogeneous sensors, and multimodal object appearance~\citep{pose2id,ye-reid,mpreid,mdreid,stmi,demo,reid-moe,mambapro,idea}. \papername{} differs by using tracking and re-identification as components of long-term robot memory, linking repeated observations of people and object instances to spatial observations and interaction histories for ownership-conditioned navigation.\\
\textbf{Object Ownership.} Recent work has studied social relations and personalization cues for robotic assistance.
Early systems model owner--object relations with probabilistic inference, visual cues, or demonstrations~\citep{IJAC-2019-07-140,ahfe2023interactiveOwnership}. 
Recent personalized search and navigation methods use dialogue, user feedback, open-vocabulary grounding, or image/language-specified object instances~\citep{dai2023think,findthis2023,barsellotti2024pin,khanna2024goatbench}. 
Other methods localize personal objects from limited examples or on-device visual matching~\citep{kim2024pga,paramonov2024swissdino}. \papername{} infers owner–object associations from long-term robot observations and uses them to decide where to search for or navigate to objects associated with a specific person.
\\
\textbf{Human--Object Interaction (HOI).} HOI recognition identifies how people interact with objects. 
Recent work extends HOI beyond closed-set triplets toward open-vocabulary and open-world recognition with vision-language or multimodal language models~\citep{open-voc-hoi,hola,no-more-sibling,tonini2026towards}. 
Related embodied work uses interaction reasoning for long-horizon hand-object manipulation from language~\citep{openhoi}. In contrast, \papername{} stores detected interactions with person, object, place, and time information, and uses these histories as evidence for ownership-conditioned navigation.

\begin{figure}[t]
    \centering
    \includegraphics[width=1\textwidth]{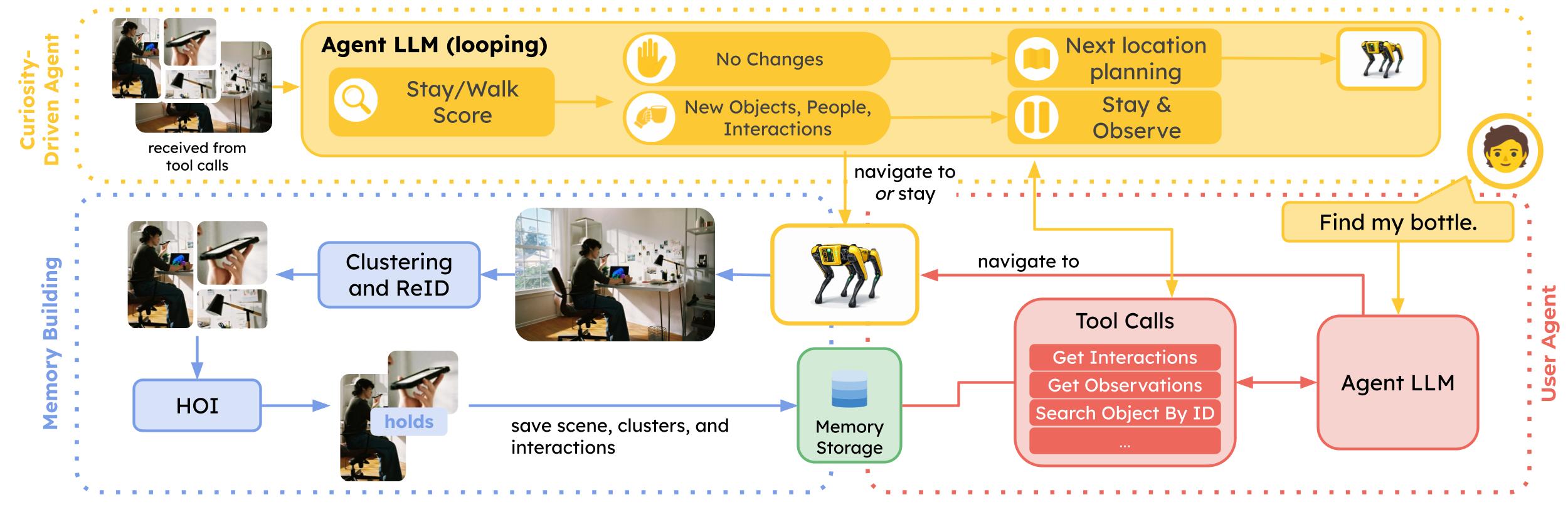}
    \vspace{-4mm}
    \caption{Our framework integrates curiosity-driven data collection with persistent spatial memory and ownership-conditioned reasoning. A curiosity-driven agent (top) decides when to stay and observe or move to more informative locations, the memory module (blue) clusters repeated person and object detections and stores human–object interactions, while a user agent (red) retrieves this evidence to infer ownership and execute personalized commands such as \textit{Find my bottle}.
    }
    \label{fig:overview}
    \vspace{-4mm}
\end{figure}%

\section{Method}
\label{sec:method}
\begin{figure}[t]
    \centering
    \includegraphics[width=1\textwidth]{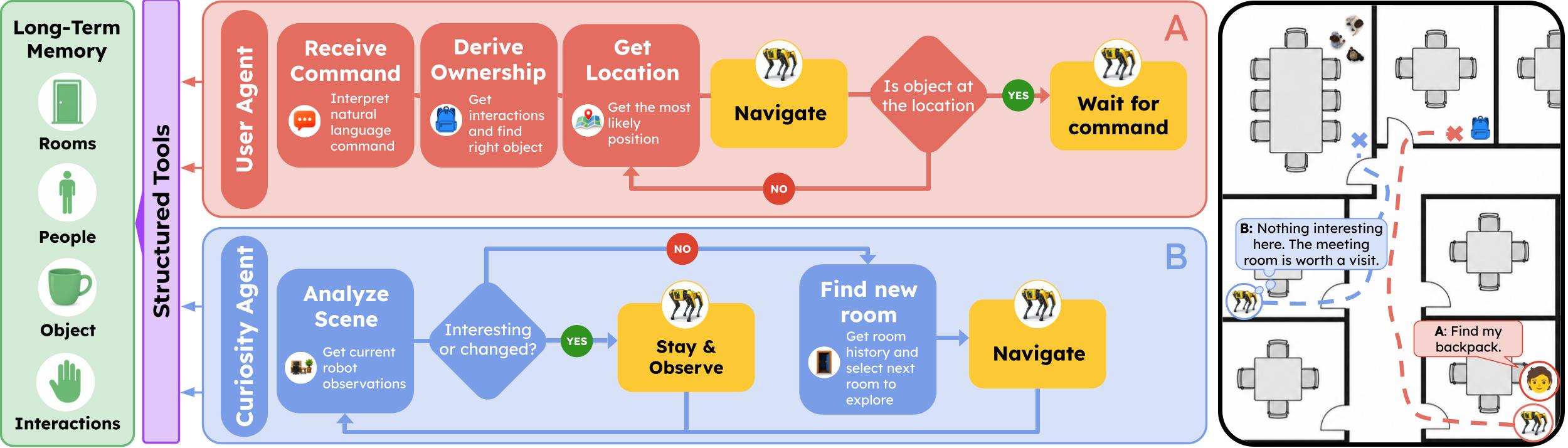}
    \vspace{-4mm}
    \caption{Agent-based reasoning in \papername{}: we use two  language agents to operate on a shared long-term spatial memory. A user agent (red) retrieves people, objects, locations, and interaction histories to infer ownership and generate navigation goals, while a curiosity-driven exploration agent (blue) decides where to go next to collect ownership-relevant evidence and refresh stale observations.}
    \label{fig:agents}
    \vspace{-4mm}
\end{figure}

The key contribution of \papername{} is ownership inference: enabling mobile robots to reason about which objects belong to which people from situated, long-term observations (Fig.~\ref{fig:overview}). During operation, the robot grounds visual observations in a metric map and stores them in a spatial memory (Sec.~\ref{sec:persistent_memory}). The system's reasoning is driven by two specialized language agents (Fig.~\ref{fig:agents}). First, a \textit{user agent} (Sec.~\ref{sec:ownership_inference}) queries this memory to resolve entities, infer ownership relations, answer ownership-related questions, and issue navigation commands. Second, a \textit{curiosity agent}~(Sec.~\ref{sec:curiosity_data_collection}) actively expands and refreshes the memory through curiosity-driven data collection.

\subsection{Persistent Person and Object Memory}
\label{sec:persistent_memory}

A persistent memory links repeated detections of the same people and object instances over time. This association is necessary because ownership evidence is accumulated across multiple observations. 
At each observation step, the robot detects visible people and objects in its RGB-D camera stream using YOLO11m-seg \cite{yolo11_ultralytics}, selected based on the results of an ablation study reported in App.~\ref{app:object_detection_ablation}.
We then assign these observations to persistent clusters with a pipeline that consists of three stages:  
First, the system performs \textit{short-term association} by preserving identity through BoTSort \citep{aharon2022bot} track continuity. If the track ID is lost, a short-term spatial cache reassigns nearby same-class detections to the previous cluster. This stage handles frame-to-frame continuity and brief detector dropouts.
Second, we perform \textit{nearest-neighbor retrieval} based on embeddings of people and objects. Specifically, we represent people with {OSNet $\times$1.0} body-appearance embeddings and objects with {ConvNeXt-Tiny}~\cite{convnext} visual embeddings. These feature extractors were selected based on ablation studies comparing several embedding models and input preprocessing variants. For person re-identification, OSNet provided the strongest performance among the off-the-shelf models. We fine-tuned it on our data to further improve its clustering performance. For object re-identification, ConvNeXt-Tiny achieved the highest F1 score among the evaluated feature extractors (App.~\ref{app:reid_ablation}, App.~\ref{app:object-reid-clustering}).
For the current observation embedding $z_t$, we retrieve the $k$ nearest stored observations across all clusters using cosine similarity. We then group these neighbors by their assigned cluster $k$ and compute a score 
\begin{equation} s_k = w \cdot \bar{s}_k + (1-w) \cdot s_k^{\max}, \end{equation} where $\bar{s}_k$ is the average cosine similarity to cluster $k$'s neighbors, $s_k^{\max}$ is the best similarity among them, and $w$ controls the trade-off between the average and the peak similarity. The best cluster is accepted only if it satisfies the embedding similarity threshold $\tau_{\text{emb}}$, receives at least $h_{\min}$ neighbor hits, and has a margin of at least $\delta$ over the runner-up. Appendix~\ref{app:clustering} lists the clustering parameters. 
Third, for predefined static classes, such as chairs and tables, the system performs \textit{position-based matching}. If an object of the same class is observed again at approximately the same location, the observation is assigned to the existing cluster. 
This stage helps separate visually similar static objects by location.

Online association can fragment one physical entity across clusters due to viewpoint changes, occlusion, or embedding noise. 
The system therefore periodically compares same-class cluster centroids and merges clusters when their accumulated appearance evidence is sufficiently similar.
Person identity is estimated from both face and body appearance. 
The association pipeline is extended with InsightFace embeddings~\cite{deng2018arcface} for face-visible detections.
Face embeddings provide high-specificity matches, while body embeddings preserve continuity when the face is not captured.

\subsection{Agent-Based Ownership Inference and Navigation}
\label{sec:ownership_inference}

We formulate ownership as a latent, dynamically inferred relation between a person cluster $p$ and an object-instance cluster $o$. Unlike explicit labels, this relation is not stored as a fixed database attribute. Instead, \papername{} infers it at query time from the robot's accumulated memory of observations and human--object interactions. This design allows ownership estimates to update when new evidence is collected, ambiguous histories become clearer, or ownership changes.

\textbf{Ownership Inference.} Given a query $q$, such as \emph{Who owns the mug?} or \emph{Go to Simba's mug}, the system first resolves the referenced entities using structured memory tools (see Appendix~\ref{app:agent_tools}). Let $\mathcal{P}$ denote the set of persistent person clusters and $\mathcal{O}$ the set of persistent object-instance clusters. For each candidate
person--object pair $(p,o) \in \mathcal{P} \times \mathcal{O}$, \papername{} retrieves an ownership evidence set
\begin{equation}
\mathcal{E}_{p,o}
=
\{ e_i = (p_i, o_i, a_i, \ell_i, t_i, c_i) \},
\end{equation}
where each event records the involved person and object clusters, the detected interaction type $a_i$, the map location $\ell_i$, timestamp $t_i$, and an optional caption $c_i$. 
Interactions are inferred from detected person--object bounding-box pairs using a VLM, inspired by recent HOI pipelines~\citep{open-voc-hoi}.
These events capture direct human--object interactions, repeated person and object co-occurrences, and their observed locations. Direct physical interactions, such as drinking, using, or picking up, provide stronger ownership evidence than holding or weak co-presence.

A tool-using language agent reasons over the retrieved evidence through a constrained multi-turn retrieval loop. Using structured tool calls, the agent retrieves candidate entities and evidence records and must define the ownership state: \emph{owned by person $p$}, \emph{shared}, or \emph{unknown}. It is instructed to weigh evidence according to interaction specificity, repetition, recency, and consistency across locations. Repeated direct interactions by the same person increase confidence in ownership, while a single weak interaction or mere spatial co-presence is insufficient. Evidence distributed across people is labeled as \textit{shared}, while weak or conflicting evidence is labeled as \textit{unknown}.

\textbf{Ownership-conditioned Navigation.} The inferred relation is used to select an object instance and map it to a reachable navigation goal. Detections are grounded in a pre-recorded metric map using \texttt{slam\_toolbox} \citep{Macenski2021}. Each detection is assigned a 3-D position by back-projecting the median depth inside the YOLO bounding box. Let
${}^{w}\mathbf{p}_{r}(t)$ and ${}^{w}\mathbf{R}_{r}(t)$ denote the robot position and orientation in the map at time $t$, and let ${}^{r}\mathbf{p}_{obj}(t)$ denote the object position in the robot frame. The corresponding global object position is
\begin{equation}
{}^{w}\mathbf{p}_{obj}(t)
=
{}^{w}\mathbf{p}_{r}(t)
+
{}^{w}\mathbf{R}_{r}(t){}^{r}\mathbf{p}_{obj}(t).
\end{equation}
For possessive commands such as \emph{Go to Simba's
mug}, the agent first infers the relevant owner--object relation, selects the object instance most strongly associated with the named person, and retrieves its most recent reliable map position. For generic commands without an owner reference, such as \emph{Go to the blue mug}, the agent selects the most recently
observed matching instance unless additional disambiguating evidence is available.

\subsection{Curiosity-Driven Data Collection}
\label{sec:curiosity_data_collection}

Since ownership is inferred from long-term evidence rather than explicit labels, the robot must decide where to spend its sensing time to observe useful events: people interacting with objects, objects appearing in new locations, and previously stored observations becoming stale. We therefore formulate curiosity-driven data collection as an active memory-refresh problem. The robot must balance three competing goals: revisiting locations where ownership evidence is likely to occur, checking locations where new changes are expected, and refreshing under-observed areas. 

To this end, we use a language-agent-based exploration loop rather than random wandering or a fixed patrol schedule. At defined intervals, the agent observes the current room and determines whether the scene has changed compared to its recent history. Changes include variations in object counts, visible people, object locations, and human--object interactions. If the agent expects near-term ownership-relevant activity, or has recently arrived and needs more observations, it issues a \textit{stay} command. Staying allows the perception pipeline to accumulate repeated detections, improve person and object association, and capture interaction events that may provide ownership evidence. 

Conversely, if the scene remains static, the agent issues a \textit{move} command.
To select the next room, the agent queries spatial memory with three tools and scores candidate rooms using three signals:  
\textbf{(1) Historical Hotspots}~identify rooms with a high empirical rate of scene changes or human--object interactions. They are likely to contain direct ownership evidence.
\textbf{(2) Expected Change}~returns the time elapsed since the last detected change in each room. Rooms that have not changed recently but historically change often are likely to provide new evidence if revisited.
\textbf{(3) Staleness}~identifies rooms whose observations are old or sparse. Visiting stale rooms prevents the robot from relying on outdated object locations and helps update ownership-conditioned navigation targets.
 
This makes curiosity explicitly tied to the ownership inference objective: the robot preferentially collects evidence that can strengthen, revise, or disambiguate person--object relations, while
also maintaining fresh object-location estimates for navigation. 
When the robot navigates to a selected destination, the perception and memory pipeline remains active. Newly detected people, objects, object locations, and human--object interactions are grounded in the metric map, associated with existing clusters when possible, and inserted into the long-term spatial memory. Through this closed-loop process, \papername{} continuously enriches and refreshes the evidence used for ownership inference.

\section{Experiments and Results}
\label{sec:experiments}

We implement the hardware-agnostic framework in ROS~2 and evaluate it on a Boston Dynamics Spot robot~\citep{bostondynamics_spot} equipped with RGB-D perception and LiDAR-based SLAM. Navigation is executed with Nav2~\citep{macenski2020marathon2, macenski2023survey} and uses the SmacPlanner2D global planner \citep{macenski2024smac} over the 2-D occupancy grid.  
We use Qwen3-VL-4B~\citep{Bai2025Qwen3VLTR} for both agents as the backbone, which is commonly used in robotics~\cite{Zhang2026-ke,Zhang2026-jo,Huang2025-pz,Zeng2025-rw}.
The model runs on an RTX~4090. For the longest queries (2k input tokens), we observe an end-to-end latency of $\sim$2~s. We store embeddings and text rather than raw images, minimizing storage requirements to 11 KB per observation. Since embeddings can enable re-identification, we obtained consent from all subjects.

\subsection{Offline Reasoning Benchmark: Synthetic Dataset}
\label{sec:synthetic_office_benchmark}

\paragraph{Synthetic Dataset.}
We evaluate the memory-reasoning capabilities of \papername{} on a synthetic office benchmark to systematically test ownership reasoning components.
The benchmark represents one day of robot perception in a shared office (see Appendix~\ref{app:gpt_dataset}).
It contains rooms, people, objects, timestamped observations, and human--object interactions.
We generated an initial trace using an LLM and manually reviewed it for plausibility. The log was evaluated by examining the recorded activities and their temporal sequence to determine whether they were consistent with a realistic office workday. Implausible activities, sequences, or transitions were identified and corrected based on human judgment. We use this synthetic dataset because, to our knowledge, no existing dataset combines navigation coordinates with ground-truth ownership data and recurring people and objects \cite{procthor, Khanna_2024_CVPR}.
From this state, we derive $79$ natural-language questions.
These questions probe the system’s ability to reason over time and space, aggregate interaction histories, infer owner--object associations, and resolve navigation targets, helping identify the strengths and failure modes of each subcomponent.
All questions were reserved for final evaluation.

We compare \papername{} with four variants. Semantic similarity is used to evaluate the answers.
\textbf{(1)~Plain LLM} receives a flattened database snapshot.
\textbf{(2)~Rule-based} maps questions to predefined SQL queries using keywords and regular expressions.
\textbf{(3)~No interaction} runs the agent without interaction-related tools.
\textbf{(4)~Heuristic} uses recency and proximity rules.

\papername{} achieves the highest overall accuracy (Table~\ref{tab:benchmark_results}). This shows the benefit of structured tool-based memory access over flattened database prompting.\\
\textbf{Limited ownership reasoning in baselines.}
\textit{Rule-based}, \textit{Plain LLM}, and \textit{Heuristic} score $0\%$ on ownership, while \textit{No Interaction} reaches $33\%$ with high variance.
\papername{} achieves $100\%$ in this oracle-memory setting, showing that reliable ownership inference requires structured access to interaction and observation histories.\\
\textbf{Importance of interaction information.}
The \textit{No Interaction} ablation performs well on location questions, but fails on interaction and ownership questions.
This confirms that human--object interactions are necessary evidence for object-use and ownership reasoning.

\begin{table*}[t]
\centering
\caption{Results on the offline reasoning benchmark.
We report overall and category-specific accuracy over three seeds, together with the average number of tool calls and wall-clock time per query.}
\label{tab:benchmark_results}
\resizebox{\textwidth}{!}{
\begin{tabular}{l c c c c c c c c}
\toprule
\textbf{Variant} &
\textbf{Accuracy (\%)} &
\textbf{Interaction (\%)} &
\textbf{Location (\%)} &
\textbf{Lookup (\%)} &
\textbf{Ownership (\%)} &
\textbf{Navigation (\%)} &
\textbf{Func. calls} &
\textbf{Time (sec)} \\
\midrule
Plain LLM
& $37.0 \pm 0.0$
& $61.0 \pm 0.0$
& $26.0 \pm 0.0$
& $\mathbf{33.0 \pm 0.0}$
& $0.0 \pm 0.0$
& $0.0 \pm 0.0$
& $0.00 \pm 0.00$
& $4.69 \pm 0.07$ \\

Heuristic
& $21.0 \pm 0.0$
& $0.0 \pm 0.0$
& $78.0 \pm 0.0$
& $0.0 \pm 0.0$
& $0.0 \pm 0.0$
& $11.0 \pm 0.0$
& $0.25 \pm 0.00$
& $0.01 \pm 0.00$ \\

No Interaction
& $36.0 \pm 2.0$
& $6.0 \pm 3.0$
& $\mathbf{96.0 \pm 0.0}$
& $\mathbf{33.0 \pm 0.0}$
& $33.0 \pm 12.0$
& $15.0 \pm 6.0$
& $4.60 \pm 0.15$
& $6.85 \pm 0.68$ \\

Rule-based
& $15.0 \pm 0.0$
& $18.0 \pm 0.0$
& $37.0 \pm 0.0$
& $0.0 \pm 0.0$
& $0.0 \pm 0.0$
& $11.0 \pm 0.0$
& $0.25 \pm 0.00$
& $\mathbf{0.01 \pm 0.00}$ \\

\midrule
\papername{}
& $\mathbf{89.0 \pm 2.0}$
& $\mathbf{94.0 \pm 0.0}$
& $\mathbf{96.0 \pm 0.0}$
& $\mathbf{33.0 \pm 0.0}$
& $\mathbf{100.0 \pm 0.0}$
& $\mathbf{89.0 \pm 0.13}$
& $2.18 \pm 0.05$
& $2.35 \pm 0.10$ \\

\bottomrule
\end{tabular}
}
\vspace{-2mm}
\end{table*}

\subsection{Video-Based Evaluation: Sitcom Dataset}
\label{sec:sitcom_benchmark}

\paragraph{Sitcom Dataset.}
To validate ownership reasoning, we evaluate \papername{} on three hours of multi-person sitcom footage.
It captures realistic scenarios with object passing, identity changes due to clothing or viewpoint shifts, and ambiguous ownership.
Existing datasets lack the combination of recurring identities, realistic interactions, and ground-truth ownership labels \cite{procthor, Khanna_2024_CVPR}.

We manually inspect the videos to identify objects and formulate 30 ownership questions.
The system processes the video end-to-end with no oracle inputs, including missed detections and identity errors from the perception pipeline.
A question is correct if the VLM identifies the owner or correctly labels an object as shared.

\paragraph{Robustness to Perception Errors.}

\begin{wraptable}{r}{0.40\linewidth}
    \centering
    \vspace{-12pt}
    \caption{Robustness test on sitcom footage. Accuracy (\%) as mean $\pm$ standard deviation (10 seeds).}
    \label{tab:robustness}

    \resizebox{\linewidth}{!}{%
    \begin{tabular}{lr}
        \toprule
        \textbf{Robustness Test} & \textbf{Accuracy (\%)} \\
        \midrule
        Baseline & 84.33$\pm$2.75 \\
        Detection Dropout (40\%) & 81.48$\pm$5.30 \\
        Detection Dropout (60\%) & 61.12$\pm$5.78 \\
        Detection Dropout (80\%) & 47.78$\pm$5.77 \\
        False-Positive Injection (10\%) & 78.51$\pm$5.78 \\
        False-Positive Injection (50\%) & 66.67$\pm$5.99 \\
        Face Occlusion (50\%) & 79.63$\pm$2.00 \\
        Track Fragmentation (20\%) & 63.68$\pm$6.95 \\
        Identity Swap (50\%) & 50.01$\pm$12.47 \\
        \bottomrule
    \end{tabular}%
    }
    \vspace{-8pt}
\end{wraptable}

We evaluate robustness to perception errors by perturbing the sitcom dataset observations with failure modes.
We simulate eight failure modes (Table~\ref{tab:robustness}): detection dropout (40\%, 60\%, 80\%), false-positive injection (10\%, 50\%), face occlusion (50\%), track fragmentation (20\%), and identity swaps (50\%).
Each condition is evaluated over 10 seeds.

The method maintains strong performance under moderate perception noise, with accuracy of $81.48\%$ under 40\% dropout, $78.51\%$ with 10\% false positives, and $79.63\%$ with 50\% face occlusion.
Performance degrades more substantially under severe dropout and identity swaps, reaching $61.12\%$ and $50.01\%$ respectively.
These results suggest that the approach is generally tolerant to noisy or incomplete observations, while consistent identity information is particularly important for reliable ownership matching. Details can be found in Appendix~\ref{app:robustness}.

\subsection{Real-World Experiments}

To test practical applicability, we evaluate whether \papername{} can perform real-world, ownership-conditioned navigation. 
We conduct the experiments in an office environment. 
In each experiment, the robot receives a natural-language command, such as \textit{Go to Simba's ball}. 
To succeed, the robot must identify the correct object instance from its memory and navigate to that object's most recently observed location. 
Each trial starts from a random position in the office. 
We define a trial as successful if the robot reaches the navigation goal, stops, and finishes within a $1\,\mathrm{m}$ radius of the target. We evaluate five distinct navigation tasks with 10 runs each. 
Table~\ref{tab:real_world_tasks} summarizes the tested capabilities.

\begin{table*}[t]
\centering
\scriptsize
\setlength{\tabcolsep}{3pt}
\renewcommand{\arraystretch}{1}
\caption{Real-world navigation tasks and evaluated capabilities. Accuracy is measured over 10 runs per task. A “$+$” denotes a capability that is required for solving the task, whereas “$-$” denotes a capability that is not involved in the task.
Details are provided in Appendix~\ref{app:real-world-tasks}.}\label{tab:real_world_tasks}
\scalebox{0.95}{
\begin{tabular}{p{0.26\linewidth} c c c c c c c c}
\toprule
\textbf{Task Type} &
\textbf{Ownership} &
\makecell{\textbf{Person}\\\textbf{Distinction}} &
\makecell{\textbf{Person}\\\textbf{ReID}} &
\makecell{\textbf{Object}\\\textbf{ReID}} &
\makecell{\textbf{Long-Term}\\\textbf{Tracking}} &
\makecell{\textbf{Visual}\\\textbf{Ambiguity}} &
\makecell{\makecell{\textbf{Reachability}\\\textbf{Reasoning}}} & 
\textbf{Accuracy (\%)}\\
\midrule
Room Navigation
& $-$ & $-$ & $-$ & $-$ & $-$ & $-$ & $+$ & 100 \\

Object Relocation
& $+$ & $-$ & $+$ & $+$ & $+$ & $-$ & $-$ & 90 \\

Owner-based Distinction
& $+$ & $+$ & $+$ & $+$ & $+$ & $-$ & $-$ & 100\\

Visual Ambiguity
& $+$ & $+$ & $-$ & $-$ & $-$ & $+$ & $-$ & 90 \\

Ownership-Preserving Update
& $+$ & $+$ & $-$ & $+$ & $+$ & $+$ & $-$ & 90 \\

\bottomrule
\end{tabular}
}
\end{table*}

\begin{figure}[t]
    \centering
    \includegraphics[width=1\textwidth]{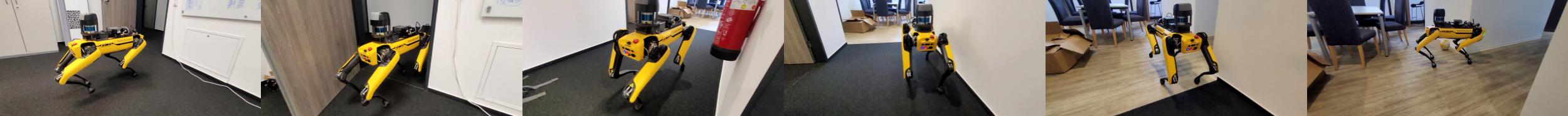}
    \includegraphics[width=1\textwidth]{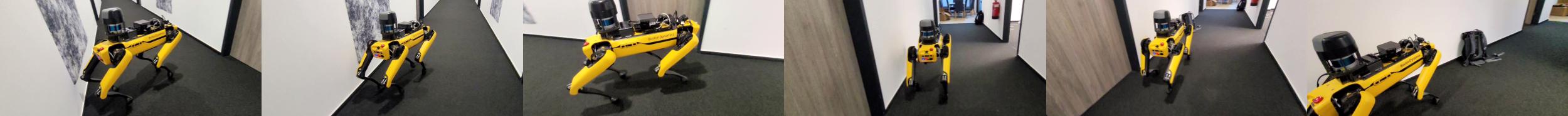}
    \includegraphics[width=1\textwidth]{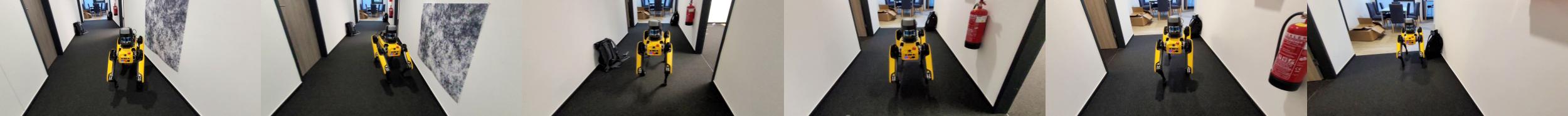}
    \vspace{-4mm}
    \caption{Three object ownership inference examples. 1) Object Relocation: Find Simba's ball (top), 2) Owner-based Distinction: Find Nemo's backpack (middle), 3) Visual Ambiguity: Find Simba's backpack (bottom). }
    \label{fig:real-world}
    \vspace{-6mm}
\end{figure}

\textbf{Results.} Table~\ref{tab:real_world_tasks} and Figure~\ref{fig:real-world} show that \papername{} successfully completes most real-world navigation trials. 
The system achieves a success rate of at least $90\%$ in every scenario, and reaches $100\%$ for \textit{Room Navigation} and \textit{Owner-based Distinction}.
This indicates that inferred ownership relations can be used for  navigation commands in a real office environment.\\
\textbf{Disambiguation and location updating.}
\textit{Visual Ambiguity} tasks show that \papername{} can disambiguate visually similar objects using interaction-derived ownership evidence. 
The \textit{Ownership-Preserving Update} task further shows that the system preserves the ownership association while updating the target location when the object is later observed elsewhere.

\subsection{Curiosity-Driven Navigation and Exploration}
\label{sec:curiosity_exploration}

We evaluated the curiosity agent on 15 synthetic office logs spanning diverse environmental conditions and change distributions. Synthetic data is necessary because existing datasets do not provide long-term, multi-person interactions with recurring ownership evidence and controlled navigation \cite{DBLP:journals/corr/abs-2110-07058, yang2025egolifeegocentriclifeassistant, kant2022housekeep}.
We set the LLM temperature to 0 to reduce sampling variability.
Each log contains 183 samples for each room, visible people, visible objects, and a scene description. 
The task is to detect changes. A change is any difference between two consecutive observations of a room: a person, object, or room fixture appearing or disappearing, or a change in what people are doing. Per-dataset details and results are presented in Appendix~\ref{app:curiosity-driven-evaluation}.

\begin{wraptable}{r}{0.50\linewidth}
\centering
\vspace{-12pt}
\caption{Navigation strategy comparison across all logs. \textit{Moves} denotes room-to-room relocations, \textit{Observed}  denotes change events that occurred while the agent was in the room, and \textit{Yield} is observed changes per move. \textit{Recall} and \textit{Precision} are measured over the agent's per-visit change judgements. \textit{Best} counts logs on which a strategy attains the highest F1 among those listed. All values are means over 15 logs × 10 seeds.}
\label{tab:nav_ablation}
    \centering
    \vspace{0.0cm}
    \resizebox{\linewidth}{!}{%
    \setlength{\tabcolsep}{2pt}
    \begin{tabular}{l c c c c c c c}
    \toprule
        Strategy        & F1     & Recall & Prec.  & Yield & Moves & Observed & Best \\ \midrule
Random          & $65.0\%$ & $88.2\%$ & $52.0\%$ & $0.36$  & $182$ & $65$ & $0/15$ \\ 
Round-Robin     & $67.9\%$ & $82.9\%$ & $57.8\%$ & $0.65$  & $91$  & $59$ & $0/15$ \\ 
Frequency       & $69.9\%$ & $86.2\%$ & $59.5\%$ & $0.50$  & $146$ & $\mathbf{73}$ & $0/15$ \\ 
Staleness       & $63.1\%$ & $\mathbf{90.6\%}$ & $48.7\%$ & $0.36$ & $182$ & $66$ & $0/15$ \\ 
Active Mapping  & $63.2\%$ & $90.5\%$ & $48.8\%$ & $0.36$  & $182$ & $65$ & $0/15$ \\ 
Greedy Hazard   & $76.5\%$ & $83.3\%$ & $71.3\%$ & $0.64$  & $108$ & $69$ & $0/15$ \\ 
Entropy         & $80.1\%$ & $70.7\%$ & $\mathbf{93.3\%}$ & $\mathbf{5.15}$ & $\mathbf{14}$ & $72$ & $1/15$ \\ 
Beta-Entropy    & $72.0\%$ & $83.4\%$ & $63.8\%$ & $0.38$  & $182$ & $69$ & $0/15$ \\ 
\midrule 
\textbf{COOL}   & $\mathbf{85.0\%}$ & $85.0\%$ & $86.0\%$ & $2.11$ & $30$ & $63$ & $\mathbf{14/15}$ \\
    \bottomrule
    \end{tabular}%
    }
\vspace{-2mm}
\end{wraptable}

\textbf{Curiosity evaluation.}
In this experiment, the agent detects scene changes and decides whether to remain in its current room or relocate. Recall is the fraction of changes detected among those occurring while the agent is present. Precision is the fraction of reported changes that are correct. Detection yield is the number of observed changes per relocation. Change-detection F1 measures whether a room changed since the agent’s previous observation, evaluated only at observed steps.
Baseline strategies select rooms using observation statistics such as visit recency, change frequency, or estimated change probability. In contrast, \papername{} uses a VLM to determine whether continued observation is warranted.

Table~\ref{tab:nav_ablation} shows that \papername{} achieves the highest F1 score of 85.0\% and is the only strategy with both precision and recall above 80\%. Sweeping strategies reach high recall but poor precision (90.6\%/48.7\%), since they visit constantly and most visits have nothing to report, whereas Entropy shows the opposite pattern (93.3\%/70.7\%). \papername{} efficiently prioritizes active rooms, requiring only 30 moves compared to 182 for sweeping strategies and 91 for Round-Robin, while observing 2.11 changes per move versus 0.36–0.65 for coverage-driven policies. The advantage further increases with world size, reaching 85.4\% F1 on 24-room logs versus 78.1\% for the strongest baseline on those logs.

\textbf{Navigator ablation.} \papername{} follows the top-ranked destination in over 99\% of relocations, but its navigator additionally determines when to relocate. Replacing this relocation policy with unconditional score argmax increases moves from $30$ to $105$ and reduces F1 from $85.0\%$ to $76.4\%$, while observing fewer interaction steps per move ($1.27$ vs.\ $3.52$). The argmax variant does leave fewer changes unobserved in absolute terms (107 vs. 400 per log) at 3.5× the movement cost, so in this configuration the navigator's contribution is deciding when to relocate under a movement budget rather than where to go.

\begin{wrapfigure}{r}{0.25\textwidth}
    \centering
    \vspace{-6.0mm}
    \includegraphics[width=\linewidth]{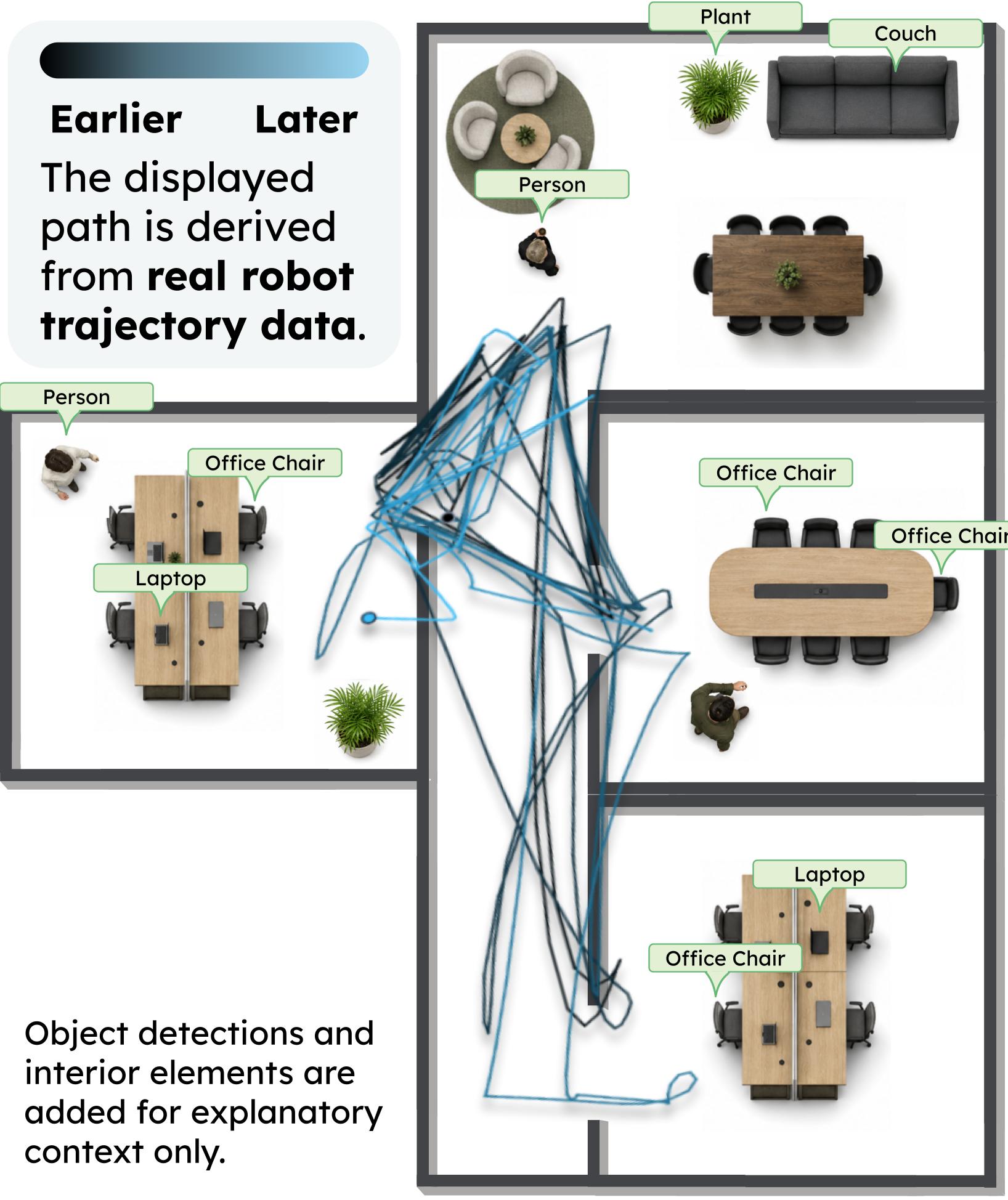}
    \vspace{-2mm}
    \caption{Real-world curiosity-driven exploration on the Spot robot. Trajectories over a 4.5-hour deployment in a four-room office environment are shown.}
    \label{fig:real-world-curiosity}
    \vspace{-14mm}
\end{wrapfigure}

\textbf{Trapping.} Entropy attains the highest precision of any strategy (93.3\%) with only 14 moves and the highest yield (5.15), but the lowest recall (70.7\% vs 85.0\% for \papername{}, Table~\ref{tab:nav_ablation}). It is accurate about the single room it occupies and blind to the rest of the building, leaving twice as many distinct change episodes unobserved as COOL (818 vs 400 per log). Trapping is an efficient way to exploit known activity, whereas COOL distributes its observations across the environment.

\textbf{Real-world run.}
We deployed the curiosity-driven agent on the robot in a four-room office environment with a kitchen and three offices. Figure~\ref{fig:real-world-curiosity} shows the resulting trajectories, where the color gradient indicates time. Across 4.5 hours of exploration, the robot autonomously navigated the environment, selecting when to move between rooms and when to remain in place, with charging breaks between runs. The robot spent more time in rooms with observed human activity, suggesting that the curiosity policy directed data collection toward interaction-rich areas.

\section{Limitations}
\label{sec:limitations}

\papername{} infers person--object ownership relations from observed behavior rather than explicit ownership labels. This enables autonomous ownership learning, but ownership remains uncertain when evidence is sparse or ambiguous, for example when objects are borrowed, moved, or only briefly handled. In such cases, the system can report ownership as unknown or shared, but additional cues such as clarification questions, visible name labels, or explicit user confirmations may be needed.
The system stores long-term observations of people, objects, locations, and interactions, which raises privacy and social acceptability concerns. Deployments in shared spaces should therefore include clear consent procedures, data minimization, access control, and retention policies.
Finally, our real-world evaluation is limited to a single office environment with a small number of people, objects, and ownership-conditioned navigation tasks. Broader studies across more environments, longer deployments, and more diverse shared-object scenarios are needed to assess generalization.

\section{Conclusion}
\label{sec:conclusion}

We presented \papername{}, a robotic framework for learning object ownership in shared human environments from long-term observations. Rather than relying on explicit labels, demonstrations, or dialogue, \papername{} infers ownership from accumulated evidence about people, object instances, and human--object interactions. This allows the system to adapt to ownership changes and incorporate new evidence.
It combines persistent spatial memory with two agents for data collection and task execution. The curiosity-driven agent autonomously explores the environment to refresh stale or incomplete memory and gather ownership-relevant evidence. The user agent interprets ownership-conditioned requests by grounding them in spatial memory, inferring the relevant owner--object relation, and directing the robot toward the inferred target location.
Our evaluations show that interaction memory, ownership-based navigation, and active observation jointly support ownership-aware assistance. On a physical robot, \papername{} improved ownership reasoning, resolved personalized search requests, collected relevant evidence more efficiently, and handled visually similar objects and changing object locations in an office environment. This suggests that persistent, interaction-aware memory and active observation can support personalized robotic assistance. Future work will explore interactive clarification, broader multi-user deployments, and mobile manipulation, extending the system beyond object localization toward retrieval and handover.

\section*{Acknowledgement}
The authors thank the anonymous reviewers for their valuable
feedback. The research was conducted under the RobOdin project funded by the European Union through its Interreg Germany-Denmark funding program. We acknowledge technical support by the Embodied AI Center at Kiel University.

\bibliography{main}  

@inproceedings{zhang2022bytetrack,
  title={{ByteTrack: Multi-Object Tracking by Associating Every Detection Box}},
  author={Zhang, Yifu and Sun, Peize and Jiang, Yi and Yu, Dongdong and Weng, Fucheng and Yuan, Zehuan and Luo, Ping and Liu, Wenyu and Wang, Xinggang},
  booktitle={Proceedings of the European Conference on Computer Vision (ECCV)},
  year={2022}
}

@article{torchreid,
  title={{Torchreid: A Library for Deep Learning Person Re-Identification in Pytorch}},
  author={Zhou, Kaiyang and Xiang, Tao},
  journal={arXiv preprint arXiv:1910.10093},
  year={2019}
}

@inproceedings{zhou2019osnet,
  title={{Omni-Scale Feature Learning for Person Re-Identification}},
  author={Zhou, Kaiyang and Yang, Yongxin and Cavallaro, Andrea and Xiang, Tao},
  booktitle={ICCV},
  year={2019}
}

@article{zhou2021osnet,
  title={{Learning Generalisable Omni-Scale Representations for Person Re-Identification}},
  author={Zhou, Kaiyang and Yang, Yongxin and Cavallaro, Andrea and Xiang, Tao},
  journal={TPAMI},
  year={2021}
}

@misc{tonini2026towards,
  title         = {{Towards Unconstrained Human-Object Interaction}},
  author        = {Tonini, Francesco and Conti, Alessandro and Vaquero, Lorenzo and Beyan, Cigdem and Ricci, Elisa},
  year          = {2026},
  eprint        = {2604.14069},
  archivePrefix = {arXiv},
  primaryClass  = {cs.CV},
  note          = {arXiv preprint arXiv:2604.14069}
}

@article{dai2023think,
  title={{Think, act, and ask: Open-world interactive personalized robot navigation}},
  author={Dai, Yinpei and Peng, Run and Li, Sikai and Chai, Joyce},
  journal={ICRA},
  year={2024}
}

@article{IJAC-2019-07-140,
author = {Wu, Hao and Chen, Zhao-Wei and Tian, Guo-Hui and Ma, Qing and Jiao, Meng-Lin},
title = {{Item Ownership Relationship Semantic Learning Strategy for Personalized Service Robot}},
year = {2020},
issue_date = {Jun 2020},
publisher = {Springer-Verlag},
address = {Berlin, Heidelberg},
volume = {17},
number = {3},
issn = {1476-8186},
journal = {Int. J. Autom. Comput.},
month = jun,
pages = {390–402},
numpages = {13}
}

@ARTICLE{efstrongsort,
  author={Khalil, Miar Mamdouh and Saleh, Sherine Nagy and Tawfik, Noha S. and Elagamy, Mazen Nabil},
  journal={IEEE Access}, 
  title={{EF-StrongSORT: An Enhanced Feature StrongSORT Model for Multi-Object Tracking}}, 
  year={2025},
  volume={13},
  number={},
  pages={53608-53620},
}

@inproceedings{
hola,
title={{HOLa: Zero-Shot HOI Detection with Low-Rank Decomposed VLM Feature Adaptation}},
author={Lei, Qinqian and Wang, Bo and Robby T., Tan},
booktitle={Proceedings of the IEEE/CVF international conference on computer vision},
year={2025}
}

@InProceedings{open-voc-hoi,
    author    = {Lei, Ting and Yin, Shaofeng and Chen, Qingchao and Peng, Yuxin and Liu, Yang},
    title     = {{Open-Vocabulary HOI Detection with Interaction-aware Prompt and Concept Calibration}},
    booktitle = {Proceedings of the IEEE/CVF International Conference on Computer Vision (ICCV)},
    month     = {October},
    year      = {2025},
    pages     = {23945-23957}
}

@InProceedings{no-more-sibling,
    author    = {Yang, Bin and Zhang, Yulin and Zhou, Hong-Yu and Yang, Sibei},
    title     = {{No More Sibling Rivalry: Debiasing Human-Object Interaction Detection}},
    booktitle = {Proceedings of the IEEE/CVF International Conference on Computer Vision (ICCV)},
    month     = {October},
    year      = {2025},
    pages     = {22707-22717}
}

@inproceedings{
    openhoi,
    title={Open{HOI}: {Open-World Hand-Object Interaction Synthesis with Multimodal Large Language Model}},
    author={Zhenhao Zhang and Ye Shi and Lingxiao Yang and Suting Ni and Qi Ye and Jingya Wang},
    booktitle={The Thirty-ninth Annual Conference on Neural Information Processing Systems},
    year={2026},
}

@inproceedings{pose2id,
  title={{From poses to identity: Training-free person re-identification via feature centralization}},
  author={Yuan, Chao and Zhang, Guiwei and Ma, Changxiao and Zhang, Tianyi and Niu, Guanglin},
  booktitle={Proceedings of the Computer Vision and Pattern Recognition Conference},
  pages={24409--24418},
  year={2025}
}

@ARTICLE{ye-reid,
  author={Ye, Hanjing and Zhao, Jieting and Zhan, Yu and Chen, Weinan and He, Li and Zhang, Hong},
  journal={IEEE Robotics and Automation Letters}, 
  title={{Person Re-Identification for Robot Person Following With Online Continual Learning}}, 
  year={2024},
  volume={9},
  number={11},
  pages={9151-9158},
  }

@inproceedings{mdreid,
  title={{MDReID: Modality-Decoupled Learning for Any-to-Any Multi-Modal Object Re-Identification}},
  author={Yingying, Feng and Jie, Li and Jie, Hu and Yukang, Zhang and Lei, Tan and Jiayi, Ji},
  booktitle={Advances in Neural Information Processing Systems},
  year={2025}
}

@article{stmi, 
title={{STMI: Segmentation-Guided Token Modulation with Cross-Modal Hypergraph Interaction for Multi-Modal Object Re-Identification}}, 
volume={40}, 
number={14}, journal={Proceedings of the AAAI Conference on Artificial Intelligence}, author={Xu, Xingguo and Liu, Zhanyu and Zhou, Weixiang and Gao, Yuansheng and Cao, Junjie and Wang, Yuhao and Luo, Jixiang and Zhang, Dell}, year={2026}, month={Mar.}, pages={11433–11441} }

@inproceedings{idea,
  title={{IDEA: Inverted Text with Cooperative Deformable Aggregation for Multi-Modal Object Re-Identification}},
  author={Wang, Yuhao and Lv, Yongfeng and Zhang, Pingping and Lu, Huchuan},
  booktitle={Proceedings of the IEEE/CVF Conference on Computer Vision and Pattern Recognition (CVPR)},
  year={2025}
}

@article{demo, title={{DeMo: Decoupled Feature-Based Mixture of Experts for Multi-Modal Object Re-Identification}}, volume={39},  number={8}, journal={Proceedings of the AAAI Conference on Artificial Intelligence}, author={Wang, Yuhao and Liu, Yang and Zheng, Aihua and Zhang, Pingping}, year={2025}, month={Apr.}, pages={8141–8149} }

@inproceedings{mambapro,
  title={{MambaPro: Multi-Modal Object Re-Identification with Mamba Aggregation and Synergistic Prompt}},
  author={Wang, Yuhao and Liu, Xuehu and Yan, Tianyu and Liu, Yang and Zheng, Aihua and Zhang, Pingping and Lu, Huchuan},
  booktitle={Proceedings of the AAAI Conference on Artificial Intelligence},
  year={2025}
}

@inproceedings{
reid-moe,
title={{Multi-Modal Object Re-identification via Sparse Mixture-of-Experts}},
author={Yingying Feng and Jie Li and Chi Xie and Lei Tan and Jiayi Ji},
booktitle={Forty-second International Conference on Machine Learning},
year={2025},
}

@misc{dinov3,
  title={{DINOv3}},
  author={Sim{\'e}oni, Oriane and Vo, Huy V. and Seitzer, Maximilian and Baldassarre, Federico and Oquab, Maxime and Jose, Cijo and Khalidov, Vasil and Szafraniec, Marc and Yi, Seungeun and Ramamonjisoa, Micha{\"e}l and Massa, Francisco and Haziza, Daniel and Wehrstedt, Luca and Wang, Jianyuan and Darcet, Timoth{\'e}e and Moutakanni, Th{\'e}o and Sentana, Leonel and Roberts, Claire and Vedaldi, Andrea and Tolan, Jamie and Brandt, John and Couprie, Camille and Mairal, Julien and J{\'e}gou, Herv{\'e} and Labatut, Patrick and Bojanowski, Piotr},
  year={2025},
  eprint={2508.10104},
  archivePrefix={arXiv},
  primaryClass={cs.CV},
  note={arXiv preprint arXiv:2508.10104},
}

@Article{convnext,
  author  = {Zhuang Liu and Hanzi Mao and Chao-Yuan Wu and Christoph Feichtenhofer and Trevor Darrell and Saining Xie},
  title   = {A ConvNet for the 2020s},
  journal = {Proceedings of the IEEE/CVF Conference on Computer Vision and Pattern Recognition (CVPR)},
  year    = {2022},
}

@inproceedings{
vit,
title={{An Image is Worth 16x16 Words: Transformers for Image Recognition at Scale}},
author={Alexey Dosovitskiy and Lucas Beyer and Alexander Kolesnikov and Dirk Weissenborn and Xiaohua Zhai and Thomas Unterthiner and Mostafa Dehghani and Matthias Minderer and Georg Heigold and Sylvain Gelly and Jakob Uszkoreit and Neil Houlsby},
booktitle={International Conference on Learning Representations},
year={2021},
}

@InProceedings{efficientnetv2,
  title = 	 {{EfficientNetV2: Smaller Models and Faster Training}},
  author =       {Tan, Mingxing and Le, Quoc},
  booktitle = 	 {Proceedings of the 38th International Conference on Machine Learning},
  pages = 	 {10096--10106},
  year = 	 {2021},
  editor = 	 {Meila, Marina and Zhang, Tong},
  volume = 	 {139},
  series = 	 {Proceedings of Machine Learning Research},
  month = 	 {18--24 Jul},
  publisher =    {PMLR},
}

@article{strongsort,
  title={Strongsort: Make deepsort great again},
  author={Du, Yunhao and Zhao, Zhicheng and Song, Yang and Zhao, Yanyun and Su, Fei and Gong, Tao and Meng, Hongying},
  journal={IEEE Transactions on Multimedia},
  year={2023},
  publisher={IEEE}
}

@InProceedings{mpreid,
    author    = {Ha, Ruiyang and Jiang, Songyi and Li, Bin and Pan, Bikang and Zhu, Yihang and Zhang, Junjie and Zhu, Xiatian and Gong, Shaogang and Wang, Jingya},
    title     = {{Multi-modal Multi-platform Person Re-Identification: Benchmark and Method}},
    booktitle = {Proceedings of the IEEE/CVF International Conference on Computer Vision (ICCV)},
    month     = {October},
    year      = {2025},
    pages     = {10251-10261}
}

@inproceedings{
    saxena2025grapheqa,
    title={Graph{EQA}: {Using 3D Semantic Scene Graphs for Real-time Embodied Question Answering}},
    author={Saumya Saxena and Blake Buchanan and Chris Paxton and Peiqi Liu and Bingqing Chen and Narunas Vaskevicius and Luigi Palmieri and Jonathan Francis and Oliver Kroemer},
    booktitle={9th Annual Conference on Robot Learning},
    year={2025},
}

@inproceedings{zhang2026fasteqa,
  title={{FAST-EQA: Efficient Embodied Question Answering with Global and Local Region Relevancy}},
  author={Haochen Zhang and Nirav Savaliya and Faizan Siddiqui and Enna Sachdeva},
  booktitle = {Winter Conference on Applications of Computer Vision (WACV), 2026},
  year={2026}
}

@article{monaci2025rana,
  title={{RANa: Retrieval-Augmented Navigation}},
  author={Monaci, Gianluca and Rezende, Rafael S and Deffayet, Romain and Csurka, Gabriela and Bono, Guillaume and D{\'e}jean, Herv{\'e} and Clinchant, St{\’e}phane and Wolf, Christian},
  journal={Transactions on Machine Learning Research},
  year={2025}
}

@article{niu2026ssmgnav,
  title = {{SSMG-Nav: Enhancing Lifelong Object Navigation with Semantic Skeleton Memory Graph}},
  author = {Niu, Haochen and Zhang, Lantao and Ji, Xingwu and Ying, Rendong and Liu, Peilin and Wen, Fei},
  journal = {arXiv preprint arXiv:2603.01813},
  year = {2026}
}

@inproceedings{anwar2025remembr,
  title     = {{ReMEmbR: Building and Reasoning Over Long-Horizon Spatio-Temporal Memory for Robot Navigation}},
  author    = {Anwar, Abrar and Welsh, John and Biswas, Joydeep and Pouya, Soha and Chang, Yan},
  booktitle = {2025 IEEE International Conference on Robotics and Automation (ICRA)},
  pages     = {2838--2845},
  year      = {2025},
  publisher = {IEEE}
}

@inproceedings{
    ginting2025mindpalace,
    title={{Enter the Mind Palace: Reasoning and Planning for Long-term Active Embodied Question Answering}},
    author={Muhammad Fadhil Ginting and Dong-Ki Kim and Xiangyun Meng and Andrzej Marek Reinke and Bandi Jai Krishna and Navid Kayhani and Oriana Peltzer and David Fan and Amirreza Shaban and Sung-Kyun Kim and Mykel Kochenderfer and Ali-akbar Agha-mohammadi and Shayegan Omidshafiei},
    booktitle={9th Annual Conference on Robot Learning},
    year={2025},
}

@inproceedings{
luo2025videorag,
    title={Video-{RAG}: {Visually-aligned Retrieval-Augmented Long Video Comprehension}},
    author={Yongdong Luo and Xiawu Zheng and Guilin Li and Shukang Yin and Haojia Lin and Chaoyou Fu and Jinfa Huang and Jiayi Ji and Fei Chao and Jiebo Luo and Rongrong Ji},
    booktitle={The Thirty-ninth Annual Conference on Neural Information Processing Systems},
    year={2026},
}

@inproceedings{
    pan2025secom,
    title={{SeCom: On Memory Construction and Retrieval for Personalized Conversational Agents}},
    author={Zhuoshi Pan and Qianhui Wu and Huiqiang Jiang and Xufang Luo and Hao Cheng and Dongsheng Li and Yuqing Yang and Chin-Yew Lin and H. Vicky Zhao and Lili Qiu and Jianfeng Gao},
    booktitle={The Thirteenth International Conference on Learning Representations},
    year={2025},
}

@article{
    du2026memguide, 
    title={{MemGuide: Intent-Driven Memory Selection for Goal-Oriented Multi-Session LLM Agents}}, 
    volume={40}, 
    number={36}, 
    journal={Proceedings of the AAAI Conference on Artificial Intelligence}, 
    author={Du, Yiming and Wang, Bingbing and He, Yang and Liang, Bin and Wang, Baojun and Li, Zhongyang and Gui, Lin and Pan, Jeff Z. and Xu, Ruifeng and Wong, Kam-Fai},
    year={2026}, month={Mar.}, pages={30584–30592} 
}

@inproceedings{
    xu2025amem,
    title={{A-Mem: Agentic Memory for {LLM} Agents}},
    author={Wujiang Xu and Zujie Liang and Kai Mei and Hang Gao and Juntao Tan and Yongfeng Zhang},
    booktitle={The Thirty-ninth Annual Conference on Neural Information Processing Systems},
    year={2026},
}

@article{chhikara2025mem0,
  title   = {{Mem0: Building Production-Ready AI Agents with Scalable Long-Term Memory}},
  author={Prateek Chhikara and Dev Khant and Saket Aryan and Taranjeet Singh and Deshraj Yadav},
  journal = {arXiv preprint arXiv:2504.19413},
  year    = {2025}
}

@article{ahfe2023interactiveOwnership,
  title={{An Interactive Learning Framework for Item Ownership Relationship in Service Robots}},
  author={Yuanda Hu and Yate Ge and Tianyue Yang and Xiaohua Sun},
  journal={Human Factors in Robots, Drones and Unmanned Systems},
  year={2023},
}

@InProceedings{findthis2023,
  title = 	 {{FindThis: Language-Driven Object Disambiguation in Indoor Environments}},
  author =       {Majumdar, Arjun and Xia, Fei and Ichter, Brian and Batra, Dhruv and Guibas, Leonidas},
  booktitle = 	 {Proceedings of The 7th Conference on Robot Learning},
  pages = 	 {1335--1347},
  year = 	 {2023},
  editor = 	 {Tan, Jie and Toussaint, Marc and Darvish, Kourosh},
  volume = 	 {229},
  series = 	 {Proceedings of Machine Learning Research},
  month = 	 {06--09 Nov},
  publisher =    {PMLR},
}

@inproceedings{barsellotti2024pin,
 author = {Barsellotti, Luca and Bigazzi, Roberto and Cornia, Marcella and Baraldi, Lorenzo and Cucchiara, Rita},
 booktitle = {Advances in Neural Information Processing Systems},
 editor = {A. Globerson and L. Mackey and D. Belgrave and A. Fan and U. Paquet and J. Tomczak and C. Zhang},
 pages = {11228--11250},
 publisher = {Curran Associates, Inc.},
 title = {{Personalized Instance-based Navigation Toward User-Specific Objects in Realistic Environments}},
 volume = {37},
 year = {2024}
}

@InProceedings{khanna2024goatbench,
    author    = {Khanna, Mukul and Ramrakhya, Ram and Chhablani, Gunjan and Yenamandra, Sriram and Gervet, Theophile and Chang, Matthew and Kira, Zsolt and Chaplot, Devendra Singh and Batra, Dhruv and Mottaghi, Roozbeh},
    title     = {{GOAT-Bench: A Benchmark for Multi-Modal Lifelong Navigation}},
    booktitle = {Proceedings of the IEEE/CVF Conference on Computer Vision and Pattern Recognition (CVPR)},
    month     = {June},
    year      = {2024},
    pages     = {16373-16383}
}

@INPROCEEDINGS{kim2024pga,
  author={Kim, Junghyun and Kang, Gi-Cheon and Kim, Jaein and Yang, Seoyun and Jung, Minjoon and Zhang, Byoung-Tak},
  booktitle={2024 IEEE/RSJ International Conference on Intelligent Robots and Systems (IROS)}, 
  title={{PGA: Personalizing Grasping Agents with Single Human-Robot Interaction}}, 
  year={2024},
  volume={},
  number={},
  pages={9277-9284},
}

@article{paramonov2024swissdino,
    title={{Swiss DINO: Efficient and Versatile Vision Framework for On-device Personal Object Search}},
    author={Paramonov, Kirill and Zhong, Jia-Xing and Michieli, Umberto and Moon, Jijoong and Ozay, Mete},
    journal={IROS},
    year={2024}
}

@INPROCEEDINGS{deng2018arcface,
  author={Deng, Jiankang and Guo, Jia and Xue, Niannan and Zafeiriou, Stefanos},
  booktitle={2019 IEEE/CVF Conference on Computer Vision and Pattern Recognition (CVPR)}, 
  title={{ArcFace: Additive Angular Margin Loss for Deep Face Recognition}}, 
  year={2019},
  volume={},
  number={},
  pages={4685-4694}
  }

@article{Bai2025Qwen3VLTR,
      title={{Qwen3-VL Technical Report}}, 
      author={Shuai Bai and Yuxuan Cai and Ruizhe Chen and Keqin Chen and Xionghui Chen and Zesen Cheng and Lianghao Deng and Wei Ding and Chang Gao and Chunjiang Ge and Wenbin Ge and Zhifang Guo and Qidong Huang and Jie Huang and Fei Huang and Binyuan Hui and Shutong Jiang and Zhaohai Li and Mingsheng Li and Mei Li and Kaixin Li and Zicheng Lin and Junyang Lin and Xuejing Liu and Jiawei Liu and Chenglong Liu and Yang Liu and Dayiheng Liu and Shixuan Liu and Dunjie Lu and Ruilin Luo and Chenxu Lv and Rui Men and Lingchen Meng and Xuancheng Ren and Xingzhang Ren and Sibo Song and Yuchong Sun and Jun Tang and Jianhong Tu and Jianqiang Wan and Peng Wang and Pengfei Wang and Qiuyue Wang and Yuxuan Wang and Tianbao Xie and Yiheng Xu and Haiyang Xu and Jin Xu and Zhibo Yang and Mingkun Yang and Jianxin Yang and An Yang and Bowen Yu and Fei Zhang and Hang Zhang and Xi Zhang and Bo Zheng and Humen Zhong and Jingren Zhou and Fan Zhou and Jing Zhou and Yuanzhi Zhu and Ke Zhu},
	  journal={arXiv preprint arXiv:2511.21631},
      year={2025}
}

@article{aharon2022bot,
  title={{BoT-SORT: Robust Associations Multi-Pedestrian Tracking}},
  author={Aharon, Nir and Orfaig, Roy and Bobrovsky, Ben-Zion},
  journal={arXiv preprint arXiv:2206.14651},
  year={2022}
}

@misc{yolo11_ultralytics,
  author = {Glenn Jocher and Jing Qiu},
  title = {Ultralytics YOLO11},
  version = {11.0.0},
  year = {2024},
  url = {https://github.com/ultralytics/ultralytics},
  license = {AGPL-3.0}
}

@article{Zhang2026-jo,
  title   = {{Pelican-Unify 1.0: A Unified Embodied Intelligence Model for Understanding, Reasoning, Imagination and Action}},
  author  = {Zhang, Yi and Chen, Yinda and Liu, Che and Ding, Zeyuan and Xu, Jin and Zou, Shilong and Liao, Junwei and Hu, Jiayu and Ren, Xiancong and Zhang, Xiaopeng and Liu, Yechi and Shi, Haoyuan and Tang, Zecong and Sun, Haosong and Cui, Renwen and Wu, Kuishu and Liu, Wenhai and Xu, Yang and Zhang, Yingji and Wang, Yidong and Hu, Senkang and Lu, Jinpeng and Chan, Nga Teng and Wu, Yechen and Liu, Zeting and Hou, Xianzhou and Dai, Yong and Tang, Jian and Ju, Xiaozhu},
  journal = {arXiv preprint arXiv:2605.15153},
  year    = {2026}
}

@article{Zhang2026-ke,
  title   = {{Enhancing Lightweight Vision Language Models through Group Competitive Learning for Socially Compliant Navigation}},
      author={Xinyu Zhang and Atsushi Konno and Toshihiko Yamasaki and Ling Xiao},
  journal = {arXiv preprint arXiv:2603.11447},
  year    = {2026}
}

@article{Huang2025-pz,
  title   = {{NoTVLA}: {Semantics-Preserving Robot Adaptation via Narrative Action Interfaces}},
  author  = {Huang, Zheng and Liu, Mingyu and Lin, Xiaoyi and Zhu, Muzhi and Zhao, Canyu and Du, Zongze and Lin, Ye and Li, Xiaoman and Jia, Yiduo and Zhong, Hao and Chen, Hao and Shen, Chunhua},
  journal = {arXiv preprint arXiv:2510.03895},
  year    = {2025}
}

@article{Zeng2025-rw,
  title={Diagnose, Correct, and Learn from Manipulation Failures via Visual Symbols},
  author={Zeng, Xianchao and Zhou, Xinyu and Li, Youcheng and Shi, Jiayou and Li, Tianle and Chen, Liangming and Ren, Lei and Li, Yong-Lu},
  journal={arXiv preprint arXiv:2512.02787},
  year={2025}
}

@misc{bostondynamics_spot,
  author = {{Boston Dynamics}},
  title  = {{Spot}: The Agile Mobile Robot},
  year   = {2026},
  note   = {Boston Dynamics product documentation}
}

@article{macenski2024smac,
  title   = {Cost-Aware Kinematically Feasible Planning for Mobile and Surface Robotics},
  author  = {Steve Macenski and Matthew Booker and Josh Wallace and Tobias Fischer},
  journal = {IEEE Robotics and Automation Practice},
  year    = {2026}
}

@inproceedings{macenski2020marathon2,
  title     = {The Marathon 2: A Navigation System},
  author    = {Macenski, Steve and Martín, Francisco and White, Ruffin and Ginés Clavero, Jonatan},
  year      = {2020},
  booktitle = {2020 IEEE/RSJ International Conference on Intelligent Robots and Systems (IROS)},
  url       = {https://github.com/ros-planning/navigation2}
}

@article{macenski2023survey,
  title   = {From the desks of ROS maintainers: A survey of modern \& capable mobile robotics algorithms in the robot operating system 2},
  author = {Steve Macenski and Tom Moore and David V. Lu and Alexey Merzlyakov and Michael Ferguson},
  year    = {2023},
  journal = {Robotics and Autonomous Systems}
}

@article{Macenski2021, doi = {10.21105/joss.02783}, year = {2021}, publisher = {The Open Journal}, volume = {6}, number = {61}, pages = {2783}, author = {Macenski, Steve and Jambrecic, Ivona}, title = {SLAM Toolbox: SLAM for the dynamic world}, journal = {Journal of Open Source Software} }

@InProceedings{pmlr-v139-touvron21a,
  title = 	 {Training data-efficient image transformers \& distillation through attention},
  author =       {Touvron, Hugo and Cord, Matthieu and Douze, Matthijs and Massa, Francisco and Sablayrolles, Alexandre and Jegou, Herve},
  booktitle = 	 {Proceedings of the 38th International Conference on Machine Learning},
  pages = 	 {10347--10357},
  year = 	 {2021},
  editor = 	 {Meila, Marina and Zhang, Tong},
  volume = 	 {139},
  series = 	 {Proceedings of Machine Learning Research},
  month = 	 {18--24 Jul},
  publisher =    {PMLR}
}

@inproceedings{
loshchilov2018decoupled,
title={Decoupled Weight Decay Regularization},
author={Ilya Loshchilov and Frank Hutter},
booktitle={International Conference on Learning Representations},
year={2019},
}

@INPROCEEDINGS{8578114,
  author={Wei, Longhui and Zhang, Shiliang and Gao, Wen and Tian, Qi},
  booktitle={2018 IEEE/CVF Conference on Computer Vision and Pattern Recognition}, 
  title={Person Transfer GAN to Bridge Domain Gap for Person Re-identification}, 
  year={2018},
  volume={},
  number={},
  pages={79-88},
}

@article{liu2023grounding,
  title={{Grounding dino: Marrying dino with grounded pre-training for open-set object detection}},
  author={Liu, Shilong and Zeng, Zhaoyang and Ren, Tianhe and Li, Feng and Zhang, Hao and Yang, Jie and Li, Chunyuan and Yang, Jianwei and Su, Hang and Zhu, Jun and others},
  journal={arXiv preprint arXiv:2303.05499},
  year={2023}
}

@article{minderer2024scalingopenvocabularyobjectdetection,
      title={{Scaling Open-Vocabulary Object Detection}}, 
      author={Matthias Minderer and Alexey Gritsenko and Neil Houlsby},
      year={2024},
      eprint={2306.09683},
      archivePrefix={arXiv},
      primaryClass={cs.CV},
      journal={arXiv preprint arXiv:2306.09683},
}

@InProceedings{pmlr-v139-radford21a,
  title = 	 {{Learning Transferable Visual Models From Natural Language Supervision}},
  author =       {Radford, Alec and Kim, Jong Wook and Hallacy, Chris and Ramesh, Aditya and Goh, Gabriel and Agarwal, Sandhini and Sastry, Girish and Askell, Amanda and Mishkin, Pamela and Clark, Jack and Krueger, Gretchen and Sutskever, Ilya},
  booktitle = 	 {Proceedings of the 38th International Conference on Machine Learning},
  pages = 	 {8748--8763},
  year = 	 {2021},
  editor = 	 {Meila, Marina and Zhang, Tong},
  volume = 	 {139},
  series = 	 {Proceedings of Machine Learning Research},
  month = 	 {18--24 Jul},
  publisher =    {PMLR}
}

@article { LanczosFilteringinOneandTwoDimensions,
      author = "Claude E.  Duchon",
      title = {{Lanczos Filtering in One and Two Dimensions}},
      journal = "Journal of Applied Meteorology and Climatology",
      year = "1979",
      publisher = "American Meteorological Society",
      address = "Boston MA, USA",
      volume = "18",
      number = "8",
      pages=      "1016 - 1022",
}

@inproceedings{
ravi2025sam,
title={{SAM} 2: {Segment Anything in Images and Videos}},
author={Nikhila Ravi and Valentin Gabeur and Yuan-Ting Hu and Ronghang Hu and Chaitanya Ryali and Tengyu Ma and Haitham Khedr and Roman R{\"a}dle and Chloe Rolland and Laura Gustafson and Eric Mintun and Junting Pan and Kalyan Vasudev Alwala and Nicolas Carion and Chao-Yuan Wu and Ross Girshick and Piotr Dollar and Christoph Feichtenhofer},
booktitle={The Thirteenth International Conference on Learning Representations},
year={2025},
}

@InProceedings{wang2021realesrgan,
    author    = {Xintao Wang and Liangbin Xie and Chao Dong and Ying Shan},
    title     = {Real-ESRGAN: Training Real-World Blind Super-Resolution with Pure Synthetic Data},
    booktitle = {International Conference on Computer Vision Workshops (ICCVW)},
    year      = {2021}
}

@inproceedings{liu2021Swin,
  title={{Swin Transformer: Hierarchical Vision Transformer using Shifted Windows}},
  author={Liu, Ze and Lin, Yutong and Cao, Yue and Hu, Han and Wei, Yixuan and Zhang, Zheng and Lin, Stephen and Guo, Baining},
  booktitle={Proceedings of the IEEE/CVF International Conference on Computer Vision (ICCV)},
  year={2021}
}

@inproceedings{huber2026humembr,
  title     = {HUMEMBR: Learning Human Routines for Predictive Embodied Navigation},
  author    = {Huber, Samira and Pelzer, Klaas and Nguyen, Duc M. and Xiao, Xuesu and Pirk, S{\"o}ren},
  booktitle = {IEEE/RSJ International Conference on Intelligent Robots and Systems (IROS)},
  year      = {2026},
}

@article{DBLP:journals/corr/abs-2110-07058,
  author       = {Kristen Grauman and
                  Andrew Westbury and
                  Eugene Byrne and
                  Zachary Chavis and
                  Antonino Furnari and
                  Rohit Girdhar and
                  Jackson Hamburger and
                  Hao Jiang and
                  Miao Liu and
                  Xingyu Liu and
                  Miguel Martin and
                  Tushar Nagarajan and
                  Ilija Radosavovic and
                  Santhosh Kumar Ramakrishnan and
                  Fiona Ryan and
                  Jayant Sharma and
                  Michael Wray and
                  Mengmeng Xu and
                  Eric Zhongcong Xu and
                  Chen Zhao and
                  Siddhant Bansal and
                  Dhruv Batra and
                  Vincent Cartillier and
                  Sean Crane and
                  Tien Do and
                  Morrie Doulaty and
                  Akshay Erapalli and
                  Christoph Feichtenhofer and
                  Adriano Fragomeni and
                  Qichen Fu and
                  Christian Fuegen and
                  Abrham Gebreselasie and
                  Cristina Gonz{\'{a}}lez and
                  James Hillis and
                  Xuhua Huang and
                  Yifei Huang and
                  Wenqi Jia and
                  Weslie Khoo and
                  J{\'{a}}chym Kol{\'{a}}r and
                  Satwik Kottur and
                  Anurag Kumar and
                  Federico Landini and
                  Chao Li and
                  Yanghao Li and
                  Zhenqiang Li and
                  Karttikeya Mangalam and
                  Raghava Modhugu and
                  Jonathan Munro and
                  Tullie Murrell and
                  Takumi Nishiyasu and
                  Will Price and
                  Paola Ruiz Puentes and
                  Merey Ramazanova and
                  Leda Sari and
                  Kiran K. Somasundaram and
                  Audrey Southerland and
                  Yusuke Sugano and
                  Ruijie Tao and
                  Minh Vo and
                  Yuchen Wang and
                  Xindi Wu and
                  Takuma Yagi and
                  Yunyi Zhu and
                  Pablo Arbel{\'{a}}ez and
                  David Crandall and
                  Dima Damen and
                  Giovanni Maria Farinella and
                  Bernard Ghanem and
                  Vamsi Krishna Ithapu and
                  C. V. Jawahar and
                  Hanbyul Joo and
                  Kris Kitani and
                  Haizhou Li and
                  Richard A. Newcombe and
                  Aude Oliva and
                  Hyun Soo Park and
                  James M. Rehg and
                  Yoichi Sato and
                  Jianbo Shi and
                  Mike Zheng Shou and
                  Antonio Torralba and
                  Lorenzo Torresani and
                  Mingfei Yan and
                  Jitendra Malik},
  title        = {Ego4D: Around the World in 3, 000 Hours of Egocentric Video},
  journal      = {CoRR},
  volume       = {abs/2110.07058},
  year         = {2021},
  url          = {https://arxiv.org/abs/2110.07058},
  eprinttype   = {arXiv},
  eprint       = {2110.07058},
  bibsource    = {dblp computer science bibliography, https://dblp.org}
}

@misc{yang2025egolifeegocentriclifeassistant,
      title={EgoLife: Towards Egocentric Life Assistant}, 
      author={Jingkang Yang and Shuai Liu and Hongming Guo and Yuhao Dong and Xiamengwei Zhang and Sicheng Zhang and Pengyun Wang and Zitang Zhou and Binzhu Xie and Ziyue Wang and Bei Ouyang and Zhengyu Lin and Marco Cominelli and Zhongang Cai and Yuanhan Zhang and Peiyuan Zhang and Fangzhou Hong and Joerg Widmer and Francesco Gringoli and Lei Yang and Bo Li and Ziwei Liu},
      year={2025},
      eprint={2503.03803},
      archivePrefix={arXiv},
      primaryClass={cs.CV},
      url={https://arxiv.org/abs/2503.03803}, 
}

@misc{kant2022housekeep,
            title={Housekeep: Tidying Virtual Households using Commonsense Reasoning}, 
            author={Yash Kant and Arun Ramachandran and Sriram Yenamandra and Igor Gilitschenski and Dhruv Batra and Andrew Szot and Harsh Agrawal},
            year={2022},
            eprint={2205.10712},
            archivePrefix={arXiv},
            primaryClass={cs.CV}}

@inproceedings{procthor,
  author={Matt Deitke and Eli VanderBilt and Alvaro Herrasti and
          Luca Weihs and Jordi Salvador and Kiana Ehsani and
          Winson Han and Eric Kolve and Ali Farhadi and
          Aniruddha Kembhavi and Roozbeh Mottaghi},
  title={{ProcTHOR: Large-Scale Embodied AI Using Procedural Generation}},
  booktitle={NeurIPS},
  year={2022},
  note={Outstanding Paper Award}
}

@InProceedings{Khanna_2024_CVPR,
    author    = {Khanna, Mukul and Mao, Yongsen and Jiang, Hanxiao and Haresh, Sanjay and Shacklett, Brennan and Batra, Dhruv and Clegg, Alexander and Undersander, Eric and Chang, Angel X. and Savva, Manolis},
    title     = {Habitat Synthetic Scenes Dataset (HSSD-200): An Analysis of 3D Scene Scale and Realism Tradeoffs for ObjectGoal Navigation},
    booktitle = {Proceedings of the IEEE/CVF Conference on Computer Vision and Pattern Recognition (CVPR)},
    month     = {June},
    year      = {2024},
    pages     = {16384-16393}
}
\newpage
\appendix
\FloatBarrier
\section{Object and Person Clustering}
\label{app:clustering}

The system uses a fixed set of parameters for persistent person and object memory.
These parameters control embedding-based re-identification, face matching, nearest-neighbor scoring, spatial and temporal gating, and periodic cluster consolidation.
They define when observations are linked to existing clusters, when face evidence is accepted, and when cached or spatially constrained matches are allowed.
The same values are used across all experiments unless stated otherwise.
Table~\ref{tab:clustering_params} summarizes these parameters.

\begin{table}[htbp]
  \centering
  \caption{Clustering parameters used at inference time for person/object re-identification.}
  \label{tab:clustering_params}
  \resizebox{\textwidth}{!}{%
  \begin{tabular}{@{}llll@{}}
  \toprule
  \textbf{Parameter} & \textbf{Symbol} & \textbf{Value} & \textbf{Description} \\
  \midrule

  \multicolumn{4}{l}{\textit{Embedding similarity}} \\
  Visual embedding threshold & $\tau_{\text{emb}}$ & 0.643 & Cosine similarity; runtime override (code default 0.85) \\
  Face embedding threshold & $\tau_{\text{face}}$ & 0.60 & Cosine similarity; runtime override (code default 0.68) \\
  Similarity metric & -- & Cosine & pgvector \texttt{<=>} cosine distance; $\text{sim}=1-\text{dist}$ \\
  Centroid update & -- & Arithmetic mean & SQL \texttt{AVG(oo.embedding)} over cluster observations \\
  Attribute part embeddings & -- & false & Part-level attribute embeddings are disabled \\

  \midrule
  \multicolumn{4}{l}{\textit{K-NN match scoring}} \\
  K-NN neighbors & $k$ & 12 & Neighbors considered for match scoring \\
  Average weight & $w$ & 0.65 & Weight on $\bar{s}_k$ vs.\ $s_k^{\max}$ in score \\
  Min.\ score margin & $\delta$ & 0.03 & Required gap to second-best cluster \\
  Face score margin & $\delta_{\text{face}}$ & 0.03 & Min.\ margin over second-best face match \\
  Min.\ neighbor hits & $h_{\min}$ & 1 & Min.\ neighbors that must exceed threshold \\
  Min.\ observation quality & $q_{\min}$ & 0.05 & Quality floor for observations entering ReID \\

  \midrule
  \multicolumn{4}{l}{\textit{Spatial / temporal constraints}} \\
  Max.\ ReID spatial distance & $d_{\text{max}}$ & 2.0 m & Spatial veto if observation $>2$\,m from cluster centroid \\
  Track ID TTL & $\Delta t_{\text{track}}$ & 8.0 s & Time-only; track-to-object mapping expires after 8\,s \\
  Face-track TTL & $\Delta t_{\text{face-track}}$ & 20.0 s & Face-to-track association expires after 20\,s \\
  Face-track DB lookup window & $\Delta t_{\text{face-db}}$ & 180.0 s & Window for looking up face tracks in the database \\
  Recent detection cache TTL & $\Delta t_{\text{cache}}$ & 30.0 s & Hardcoded; suppresses duplicate object creation \\
  Recent detection cache distance & $d_{\text{cache}}$ & 1.0 m & $d_{\text{max}} \times 0.5$; spatial radius for cache lookup \\
  Static object position distance & $d_{\text{static}}$ & 0.8 m & Fallback match radius for non-person static objects \\
  Position source preference & -- & depth & Preferred source for observation position estimates \\

  \midrule
  \multicolumn{4}{l}{\textit{Face detection and embedding}} \\
  Minimum face size & -- & 16 px & Minimum accepted face crop size \\
  Minimum face confidence & -- & 0.45 & Minimum confidence for face detections \\
  Maximum faces per person & -- & 1 & Maximum number of face crops used per person detection \\
  Face crop margin & -- & 0.24 & Relative margin added around face crop \\
  InsightFace model & -- & \texttt{buffalo\_l} & Face embedding model \\
  InsightFace device & -- & \texttt{cuda} & Device used for face embedding inference \\
  InsightFace detection size & -- & 640 & Input detection size for InsightFace \\

  \midrule
  \multicolumn{4}{l}{\textit{Cluster consolidation (periodic merge)}} \\
  Consolidation interval & $N_{\text{consol}}$ & 120 obs & Runs every 120 new observations \\
  Consolidation similarity threshold & $\tau_{\text{consol}}$ & 0.643 & Cosine similarity for merge candidates (code default 0.80) \\
  Min.\ observations per cluster & $n_{\min}$ & 2 & Clusters below this are skipped \\
  Max.\ pairs evaluated per run & $P_{\max}$ & 6 & Caps merge computation per consolidation pass \\
  Min.\ observation quality (consolidation) & -- & 0.05 & Same as main ReID quality floor \\

  \bottomrule
  \end{tabular}%
  }
\end{table}

\FloatBarrier
\section{Structured Database Tools}
\label{app:agent_tools}

The agents interact with the memory system through a fixed set of structured tools. Each tool defines typed inputs and outputs, allowing the agent to search tracked entities, inspect observations, query interaction histories, and resolve navigation targets. These tools also provide access to the robot's current pose, enabling decisions to be grounded in the robot's spatial context. Navigation is executed only after a target has been resolved to a reachable position on the map.

\subsection{Tool Calls for the User Agent}

The user agent uses the tools listed in Table~\ref{tab:agent_tools} to answer ownership-related navigation commands. These tools allow the agent to identify people and objects, retrieve their observation histories, inspect object locations, query interaction evidence, and execute navigation commands once a valid target has been selected.

\begin{table*}[h]
  \centering
  \scriptsize
  \setlength{\tabcolsep}{3.5pt}
  \renewcommand{\arraystretch}{1.06}
  \caption{Structured tools exposed to the user agent for object search, inspection, interaction retrieval, scene understanding, and navigation.}
  \label{tab:agent_tools}
  \resizebox{0.9\textwidth}{!}{%
  \begin{tabular}{@{}llll@{}}
  \toprule
  \textbf{Tool} & \textbf{Input} & \textbf{Output} & \textbf{Purpose}\\
  \midrule
  \multicolumn{4}{@{}l}{\textit{Discovery and search}} \\
  \texttt{search\_objects\_by\_class\_id} & class\_id, name, min\_obs, limit & object list & Find tracked objects or people by class or name \\
  \texttt{list\_objects} & class\_name (opt.), limit & object list & Enumerate all tracked objects, optionally filtered \\
  \texttt{find\_person\_by\_name} & person\_name & person entity & Resolve a named person to their latest location \\
  \midrule
  \multicolumn{4}{@{}l}{\textit{Object inspection}} \\
  \texttt{get\_object\_observations} & object\_id, time filters, limit & observation history & Retrieve the location history of one entity \\
  \texttt{get\_object\_last\_location} & object\_id & latest pose & Retrieve the most recent recorded location \\
  \texttt{get\_object\_first\_location} & object\_id & earliest pose & Retrieve the first recorded location of an object \\
  \texttt{get\_object\_summary} & object\_id & spatial summary & Summarize all places where an object has been observed \\
  \texttt{get\_object\_image} & object\_id & image data & Fetch the latest cropped object image for display \\
  \midrule
  \multicolumn{4}{@{}l}{\textit{Location and navigation}} \\
  \texttt{get\_robot\_location} & -- & current pose + room & Ground decisions relative to the robot's current position \\
  \texttt{get\_room\_navigation\_target} & room\_name & navigable pose & Resolve a room to a reachable entry point \\
  \texttt{get\_objects\_in\_room} & room\_name, limit & object list & List entities currently assigned to a room \\
  \texttt{move\_to\_position} & x, y, reason, standoff\_m, yaw & nav status & Execute navigation to a resolved map target \\
  \midrule
  \multicolumn{4}{@{}l}{\textit{Interaction history}} \\
  \texttt{get\_interaction\_events} & person, object, time, room, action, limit & event list & Retrieve interaction evidence using filters \\
  \texttt{get\_object\_interactions} & object\_id, action, limit, sort & interaction list & Retrieve the history of people interacting with an object \\
  \texttt{get\_object\_person\_interactions} & object\_id, action (opt.) & aggregated list & List people who interacted with an object, including counts \\
  \texttt{get\_person\_interaction\_summary} & person\_name, other, time, room, limit & co-participants & Rank interaction partners for a person \\
  \texttt{get\_recent\_interactions} & minutes, limit & recent events & Retrieve recent interactions with room context \\
  \midrule
  \multicolumn{4}{@{}l}{\textit{Room and scene context}} \\
  \texttt{get\_current\_scene\_description} & minutes (opt.) & scene + context & Retrieve the latest caption, objects, people, and interactions \\
  \bottomrule
  \end{tabular}
  }
\end{table*}

\subsection{Tool Calls for the Curiosity-Driven Agent}

The curiosity-driven agent uses a smaller set of tools to decide which room should be visited next. These tools provide information about room visitation history, recent interaction density, and rooms that have not been observed for a longer period. This allows the agent to prioritize navigation targets based on spatial coverage and expected information gain. Table~\ref{tab:agent_tools_cur} provides an overview.

\begin{table*}[h]
  \centering
  \scriptsize
  \setlength{\tabcolsep}{3.5pt}
  \renewcommand{\arraystretch}{1.06}
  \caption{Structured tools used by the curiosity-driven navigation strategy to select the next room to visit.}
  \label{tab:agent_tools_cur}
  \resizebox{0.9\textwidth}{!}{%
  \begin{tabular}{@{}llll@{}}
  \toprule
  \textbf{Tool} & \textbf{Input} & \textbf{Output} & \textbf{Purpose} \\
  \midrule
  \texttt{get\_room\_visit\_history} & limit & per-room statistics & Retrieve visit counts, last-visit times, and event counts per room \\
  \texttt{get\_room\_event\_density} & hours, limit & event counts & Count recent interaction events per room over a time window \\
  \texttt{get\_stale\_rooms} & threshold\_min, limit & stale room list & Identify rooms that have not been visited recently \\
  \bottomrule
  \end{tabular}
  }
\end{table*}

\FloatBarrier
\section{Offline Reasoning Benchmark}
\label{app:gpt_dataset}

The synthetic office benchmark is used to evaluate reasoning abilities over structured robot memory.
It consists of a memory trace for one simulated workday.
The trace contains rooms, people, objects, timestamped observations, and human--object interaction events.
Each object and person has a stable identity.
Each observation is associated with a time and a metric position in the map.
The question set is derived from this trace and is used only for final evaluation.
Table~\ref{tab:test_set_composition} shows the distribution of questions across evaluation categories.

\begin{table}[h]
\centering
\scriptsize
\setlength{\tabcolsep}{5pt}
\renewcommand{\arraystretch}{1.1}
\caption{Test set composition.}
\label{tab:test_set_composition}
\begin{tabular}{lr}
\toprule
\textbf{Evaluation Category} & \textbf{Questions} \\
\midrule
Interaction & 33 \\
Location & 22 \\
Navigation & 13 \\
Lookup & 6 \\
Ownership & 5 \\
\midrule
Total & 79 \\
\bottomrule
\end{tabular}
\end{table}

\paragraph{How it is used.}
The benchmark is used to evaluate the user agent.
For each question, the agent can call structured tools to access memory.
The tools expose observations, interactions, entities, rooms, and navigation targets.
The agent then returns a natural-language answer.
Answers are scored with semantic similarity against manually specified reference answers.
All questions are held out until final evaluation.

\paragraph{Question categories.}
Interaction questions ask about human--object interaction events.
Location questions ask where people or objects were observed.
Navigation questions require resolving a room, person, or object to a reachable goal.
Lookup questions require retrieving an entity by name or identifier.
Ownership questions require inferring likely ownership from interaction and observation histories.

\paragraph{Example questions.}
Examples include:
\begin{itemize}
    \item Where was \texttt{notebook\_002} at 16:20 on 2026-04-27?
    \item Who picked up \texttt{headphones\_001}?
    \item What interaction happened with \texttt{whiteboard\_001} at 09:58?
    \item Who used \texttt{notebook\_001}, and how many times?
    \item Navigate to the person who used the coffee machine most frequently.
    \item Navigate to Simba's bottle.
\end{itemize}

\FloatBarrier
\subsection{Robustness Evaluation Details}
\label{app:robustness}

To evaluate robustness to perception errors, we perturb the results returned by the
memory database at query time while leaving the underlying database unchanged.
This isolates the effect of perception errors from errors introduced during data
collection or database construction. Unless otherwise specified, perturbations are
applied independently for each evaluation seed.

Table~\ref{tab:robustness_settings} summarizes the simulated failure modes. Detection
dropout randomly removes a fraction of observation rows, simulating missed
detections. False-positive injection adds observations with incorrect identities,
simulating spurious detections or identity hallucinations. Embedding noise adds
Gaussian noise to stored feature vectors, reducing the reliability of similarity
retrieval. Track fragmentation splits observations associated with an identity into
synthetic track identifiers and removes the associated identity names, simulating
loss of temporal identity consistency.

Face occlusion removes the available face image information from selected
observations. Temporal jitter randomly perturbs timestamps within the specified
range, simulating synchronization errors. Action noise replaces interaction/action
labels with incorrect labels, while class confusion swaps object categories between
visually similar classes and removes the associated captions. Finally, systematic
identity swapping applies a consistent bijection between identities throughout the
evaluation, simulating a persistent identity-assignment error.

\begin{table}[t]
\centering
\caption{Perturbations used in the robustness evaluation.}
\label{tab:robustness_settings}
\resizebox{0.8\textwidth}{!}{%
\begin{tabular}{lll}
\toprule
\textbf{Failure Mode} & \textbf{Parameter} & \textbf{Perturbation} \\
\midrule
Detection Dropout
    & $p \in \{0.4,0.6,0.8\}$
    & Randomly remove observation rows \\
False-Positive Injection
    & $p \in \{0.1,0.5\}$
    & Add observations with incorrect identities \\
Embedding Noise
    & $\sigma=0.10$
    & Add Gaussian noise to embeddings \\
Track Fragmentation
    & $p=0.2$
    & Split tracks and remove identity names \\
Face Occlusion
    & $p=0.5$
    & Remove face image information \\
Temporal Jitter
    & $\pm300$ s
    & Perturb observation timestamps \\
Action Noise
    & $p=0.3$
    & Replace action/interaction labels \\
Class Confusion
    & $p=0.3$
    & Swap similar object classes \\
Systematic Identity Swap
    & $p=0.5$
    & Apply a consistent identity permutation \\
\bottomrule
\end{tabular}%
}
\end{table}

\FloatBarrier
\section{Real-World Tasks}
\label{app:real-world-tasks}

The real-world experiments consist of five task setups.
Each setup targets a specific capability of the robot memory system and the tool-using agent.

\subsection{Detailed Tasks}
\textbf{Task 1: Room Navigation.}
The robot receives a target room name.
The target room is selected from four different rooms.
The robot starts from multiple positions across the environment.
The agent must ground the room name to a spatial region on the map.
It must then resolve a reachable navigation goal inside the room.
The robot must navigate at least $1\,\mathrm{m}$ into the target room.

\textbf{Task 2: Object Relocation.}
The robot observes a ball at five different positions.
A person is observed with the ball at two different locations.
The agent must re-identify both the ball and the person across observations.
It must infer the correct owner--object association.
It must then query the latest known position of the ball and navigate to it.

\textbf{Task 3: Owner-based Distinction.}
The robot observes two bottles in different locations.
One person interacts with one bottle in the kitchen, another person interacts with the other bottle at his workplace.
The agent must distinguish between the two bottles and the two people.
It must infer the correct owner--object association.
It must then query the latest known position of the target bottle and navigate to it.

\textbf{Task 4: Visual Ambiguity.}
The robot observes two visually similar backpacks in different locations.
Each backpack is associated with a different person through prior interaction evidence.
The agent must distinguish between the two backpacks and the two people.
It must infer the correct owner--object association.
It must then query the latest known position of the target backpack and navigate to it.

\textbf{Task 5: Ownership-Preserving Update.}
The robot observes two visually similar backpacks in different locations.
Each backpack is associated with a different person through prior interaction evidence.
One backpack is later moved to a new location.
The agent must re-identify the moved backpack without changing its ownership association.
It must distinguish between the two backpacks and the two people.
It must infer the correct owner-object association.
It must then query the latest known position of the target backpack and navigate to it.

\subsection{Success criterion}
A run is counted as successful if the agent selects the correct target entity or room, and the robot reaches and stops at the corresponding navigation goal.
For room navigation, the robot must stop at least $1\,\mathrm{m}$ inside the target room.
For object-based tasks, the robot must navigate to the latest known position of the correct target object.
A run is counted as failed if the agent selects the wrong entity, queries an outdated location, or navigates to the wrong goal.

\subsection{Execution protocol}
The task was conducted over 10 hours of naturalistic office activity, involving 10 people and 20 objects across four rooms.
Each task is evaluated over $10$ runs.
Runs differ in the robot's starting pose and, where applicable, the target instance.
The memory state is initialized with the observations and interactions required by the task.
The agent receives only the task instruction.
It must use the structured tools to query memory and resolve the navigation target. 

\FloatBarrier
\section{Curiosity-Driven Evaluation}
\label{app:curiosity-driven-evaluation}

The synthetic office evaluation set for the curiosity-driven agent contains
15 worlds at two spatial scales: nine 6-room worlds and six 24-room worlds that evaluate the same generator under increasing scale and movement-budget pressure.
All worlds span the same three simulated days, 09:00--14:00 at 5\,min intervals,
which means $61$ steps per day and $183$ decision steps per run. Table~\ref{tab:dataset-specification} gives an overview of all the worlds.

All logs are generated by a procedural office simulator in which each person movement is explicitly represented by an intended destination, an optionally carried personal object, and a departure cue rendered in the origin room. Consequently, person–object–place consistency is ensured by construction. Each room–timestep pair is also automatically annotated with a change label and an interest label, distinguishing ownership-relevant events from transient distractors.

\begin{table}[H]
    \centering
    \caption{Dataset specification. The two floor plans differ in size, population
    and object count; every other component of the simulator is shared.}
    \label{tab:dataset-specification}
    \begin{tabular}{p{0.20\textwidth} p{0.34\textwidth} p{0.34\textwidth}}
        \toprule
        \textbf{Aspect} & \textbf{6-room floor plan} & \textbf{24-room floor plan} \\
        \midrule
        \textbf{Rooms} & 6: kitchen, storage, office 1--3, meeting room
                       & 24: 16 offices, 3 meeting rooms, kitchen, lounge, storage,
                         printer room, server room \\
        \textbf{Room types} & social, utility, office, meeting & social, utility, office, meeting \\
        \textbf{People} & 8 employees,
                          each with a home office and a role
                        & 40 employees, each with a home office and a role \\
        \textbf{Objects} & 12 personal objects with a fixed owner (laptop, red cup,
                           backpack, water bottle, notebook, pen, headphones, coffee mug,
                           tablet, jacket, yoga mat, camera), plus activity-spawned
                           temporary items
                         & 40 personal objects with a fixed owner, plus activity-spawned
                           temporary items \\
        \textbf{Fixtures} & per room, e.g.\ coffee machine, fridge, microwave (kitchen);
                            projector, whiteboard, conference table (meeting room)
                          & as left, plus printer, paper shelf, server rack \\
        \textbf{Geometry} & pairwise room distances $d \in \{1,2,3,4\}$ (adjacent
                            offices $d=1$, kitchen--office 3 $d=4$)
                          & same distance model over the larger plan \\
        \textbf{Time resolution} & 5-minute intervals, 09:00--14:00 & 5-minute intervals, 09:00--14:00 \\
        \textbf{Structure} & 3 days $\times$ 61 steps $=183$ steps; $6\times183=1098$ room observations
                           & 3 days $\times$ 61 steps $=183$ steps; $24\times183=4392$ room observations \\
        \bottomrule
    \end{tabular}
\end{table}

\paragraph{Activity model.}
At each timestep, present occupants perform activities sampled from room-specific templates, as summarized in Table~\ref{tab:activity_model}. Some activities spawn temporary items which persist with a time-to-live of $15$--$30$\,min. These dynamic elements are incorporated directly into the natural-language scene descriptions rendered for the VLM, without exposing ground-truth labels.

\begin{table}[htbp]
  \centering
  \caption{Activities and durations by room type.}
  \label{tab:activity_model}
  \begin{tabular}{lll}
    \hline
    \textbf{Room Type} & \textbf{Activity} & \textbf{Duration} \\
    \hline
    \multirow{6}{*}{Office} 
      & Works at their desk & $20$--$200$\,min \\
      & Types on their laptop & $15$--$120$\,min \\
      & Reads a document & $10$--$40$\,min \\
      & Joins a video call & $15$--$120$\,min \\
      & Writes notes & $10$--$30$\,min \\
      & Checks their phone & $5$\,min \\
    \hline
    \multirow{5}{*}{Kitchen} 
      & Makes coffee & $5$--$10$\,min \\
      & Prepares a snack & $5$--$15$\,min \\
      & Washes a mug & -- \\
      & Chats near the counter & -- \\
      & Fills a water bottle & -- \\
    \hline
    \multirow{4}{*}{Meeting Room} 
      & Sits at the table & -- \\
      & Writes on the whiteboard & -- \\
      & Points at the screen & -- \\
      & Takes notes & -- \\
    \hline
    \multirow{4}{*}{Utility Room} 
      & Organizes boxes & -- \\
      & Looks for a tool & -- \\
      & Checks inventory & -- \\
      & Sweeps the floor & -- \\
    \hline
  \end{tabular}
\end{table}

Movement is a stochastic decision layer.
At the end of an activity, a person moves with a room-type-dependent base
probability ($0.85$ kitchen, $0.70$ meeting, $0.60$ office, $0.55$ utility),
modulated by time-of-day multipliers ($\times1.8$ at lunch, $\times1.5$ in the
late afternoon, $\times0.7$ in the first half hour) and by per-world room
multipliers.
On top of this layer, every world runs a small scripted schedule of $3$ events per
day (e.g. stand-up, lunch, client presentation, architecture discussion).

\paragraph{Room priors.}
Worlds differ in how informative the marginal visit statistics are.
\emph{Hot} priors are the default (kitchen and home offices dominate), \emph{mild}
priors partially flatten the affinities, and \emph{flat} priors make every room
a priori equally attractive, so visit counts carry almost no information and the
language cues become the only usable signal.

\subsection{Change Definition}

Ground truth is computed per room by comparing consecutive observations of that
room, independently of where the robot actually is.

\paragraph{Entity-level changes.}
For each room we compare the set of people and the set of objects present at
timestep $t$ with those at $t+1$.
Any addition or removal is a change.
Its severity is classified as
\emph{major} if a person entered or left, a fixture changed, or at least two
objects changed;
\emph{minor} if exactly one object changed or only the activities changed.
\emph{Major} changes are the ownership-relevant ones.

\paragraph{Activity-level changes.}
Activity-level changes are detected by comparing the interactions between
consecutive observations, i.e.\ the set of \texttt{subject:action:target} tuples
observed in a room.
A room is labelled as changed at the activity level if this set differs between
$t$ and $t+1$. The severity of an activity-only change is \emph{minor}.
\newpage
\subsection{The 15 Evaluation Worlds}

All logs are generated by a custom office simulator that explicitly defines each person’s destination, optional personal object, and departure cue. This ensures consistency between people, objects, and rooms. The simulator also automatically labels each room at each timestep by change type (none, minor, or major) and interest (ownership-relevant or transient distractor). The released logs are generated with fixed random seeds and world layouts. Table~\ref{tab:worlds} gives an overview of the generated worlds.

\begin{table*}[h]
\centering
\caption{Evaluation worlds. Each world spans 3 simulated days,
09:00--14:00 at 5\,min intervals ($183$ decision steps/run).
\textbf{Obs.} = logged room observations; \textbf{Maj.} = ownership-relevant
scene changes; \textbf{I/D} = v2 interesting/distractor changes;
\textbf{Ep.} = scripted intent episodes; $p_{\mathrm{head}}$ =
probability that a departure cue names its destination. Occupancy priors
are \textit{hot}, \textit{mild}, or \textit{flat}. Each world is fixed;
seeds vary policy randomness only.}
\label{tab:worlds}

\footnotesize
\setlength{\tabcolsep}{3pt}
\renewcommand{\arraystretch}{1.05}

\begin{tabularx}{\textwidth}{
  l
  *{7}{r}
  c
  X
}
\toprule
\textbf{World}
& \textbf{R}
& \textbf{P}
& \textbf{O}
& \textbf{Obs.}
& \textbf{Maj.}
& \textbf{I/D}
& \textbf{Ep.}
& \textbf{$p_{\mathrm{head}}$}
& \textbf{Prior / property} \\
\midrule

\multicolumn{10}{l}{\textit{6-room worlds}}\\[-2pt]

\texttt{diffused}
& 6 & 8 & 21 & 1098 & 365 & -- & -- & -- &
hot; diffused activity \\

\texttt{regime}
& 6 & 8 & 24 & 1098 & 140 & -- & -- & -- &
hot; daily regime shift, sparse changes \\

\texttt{adversarial}
& 6 & 8 & 18 & 1098 & 296 & -- & -- & -- &
hot; moving hotspots \\

\texttt{base}
& 6 & 8 & 9 & 1098 & 234 & 121/267 & -- & -- &
hot; v1 parity under v2 generator \\

\texttt{distractor}
& 6 & 8 & 10 & 1098 & 280 & 135/273 & -- & -- &
hot; $4\times$ churn in kitchen/meeting,
mostly distractor-class \\

\texttt{intent\_cued}
& 6 & 8 & 15 & 1098 & 250 & 132/256 & 14 & -- &
hot; ownership cues precede change \\

\texttt{intent\_cued\_heading}
& 6 & 8 & 14 & 1098 & 275 & 161/245 & 19 & 0.5 &
hot; cues additionally name destination \\

\texttt{agent\_cued\_6room}
& 6 & 8 & 15 & 1098 & 255 & 137/240 & 22 & 0.6 &
mild; denser cues, reduced hotspot advantage \\

\texttt{agent\_cued\_flat\_6room}
& 6 & 8 & 11 & 1098 & 261 & 147/255 & 19 & 0.7 &
flat; cues are the only predictive signal \\

\midrule
\multicolumn{10}{l}{\textit{24-room worlds}}\\[-2pt]

\texttt{scaled\_base\_24room}
& 24 & 40 & 44 & 4392 & 1041 & 672/978 & -- & -- &
hot; scale without intent episodes \\

\texttt{scaled\_24room}
& 24 & 40 & 45 & 4392 & 1046 & 665/968 & 31 & -- &
hot; scale with intent episodes \\

\texttt{scaled\_distractor\_24room}
& 24 & 40 & 41 & 4392 & 1087 & 632/1056 & -- & -- &
hot; scale + $4\times$ distractor hotspots \\

\texttt{scaled\_heading\_24room}
& 24 & 40 & 47 & 4392 & 1055 & 660/970 & 32 & 0.5 &
hot; scale + heading cues \\

\texttt{agent\_cued\_24room}
& 24 & 40 & 45 & 4392 & 1081 & 684/970 & 44 & 0.6 &
mild; denser cues at scale \\

\texttt{agent\_cued\_flat\_24room}
& 24 & 40 & 44 & 4392 & 1107 & 695/999 & 54 & 0.7 &
flat; cue-following at scale \\

\bottomrule
\end{tabularx}

\vspace{2pt}
\begin{minipage}{\textwidth}
\scriptsize
\textit{Notes.}
R = rooms, P = people, O = objects, Obs. = R$\times$183.
I/D reports interesting/distractor changes; these labels are unavailable
in legacy v1 logs. ``--'' denotes not applicable. Occupancy priors:
\textit{hot} = kitchen/home offices dominant; \textit{mild} = partially
flattened; \textit{flat} = rooms equally attractive a priori, making
language cues the primary predictive signal. Intent episodes are
cue$\rightarrow$change events within 10--30\,min, with 10\% head fakes.
\end{minipage}
\end{table*}

\subsection{Change Distribution}

The worlds differ mainly in \emph{where} and \emph{when} changes concentrate.
Table~\ref{tab:perroom6} gives the full per-room distribution of major changes
for the nine 6-room worlds and Table~\ref{tab:perroom24} the concentration
summary for the six 24-room worlds, in both cases as 3-day totals with the Gini
coefficient of the per-room counts as a concentration index
($0$ = uniform, $1$ = all changes in one room).
Storage is naturally quiet in every hot-prior world ($2$--$16$ changes over three
days) but is never systematically suppressed, and the flat-prior worlds
deliberately remove that gradient (\texttt{agent\_cued\_flat\_6room}: Gini
$0.10$, storage as busy as an office).

\begin{table}[H]
\centering
\caption{Major changes per room, 3-day totals, 6-room worlds. \emph{Gini} is the concentration of
changes across the six rooms.}
\label{tab:perroom6}
\footnotesize
\begin{tabular}{lrrrrrrr}
\toprule
World & Kitchen & Storage & Office 1 & Office 2 & Office 3 & Meeting & Gini \\
\midrule
\texttt{default} & $96$ & $16$ & $58$ & $70$ & $43$ & $82$ & $0.24$ \\
\texttt{regime} & $32$ & $13$ & $28$ & $24$ & $25$ & $18$ & $0.15$ \\
\texttt{adversarial} & $70$ & $3$ & $54$ & $66$ & $48$ & $55$ & $0.22$ \\
\midrule
\texttt{base\_v2} & $62$ & $9$ & $50$ & $36$ & $28$ & $49$ & $0.25$ \\
\texttt{distractor} & $88$ & $2$ & $38$ & $46$ & $43$ & $63$ & $0.30$ \\
\texttt{intent\_cued} & $65$ & $10$ & $43$ & $39$ & $31$ & $62$ & $0.25$ \\
\texttt{intent\_cued\_heading} & $63$ & $10$ & $57$ & $39$ & $32$ & $74$ & $0.26$ \\
\texttt{agent\_cued\_6room} & $57$ & $16$ & $51$ & $32$ & $34$ & $65$ & $0.22$ \\
\texttt{agent\_cued\_flat\_6room} & $37$ & $35$ & $44$ & $39$ & $47$ & $59$ & $0.10$ \\
\bottomrule
\end{tabular}
\end{table}

\begin{table}[H]
\centering
\caption{Concentration of major changes in the 24-room worlds, 3-day totals. With 24 rooms no
single room dominates: the busiest room carries $7$--$14\%$ of all major changes, which is what
makes camping strategies expensive at scale.}
\label{tab:perroom24}
\footnotesize
\begin{tabular}{lrrl}
\toprule
World & Changes & Gini & Three busiest rooms (share of major changes) \\
\midrule
\texttt{scaled\_base\_24room} & $1041$ & $0.21$ & kitchen $11\%$, meeting room 1 $7\%$, office 3 $5\%$ \\
\texttt{scaled\_24room} & $1046$ & $0.27$ & kitchen $11\%$, meeting room 1 $7\%$, meeting room 3 $6\%$ \\
\texttt{scaled\_distractor\_24room} & $1087$ & $0.28$ & kitchen $14\%$, meeting room 1 $9\%$, meeting room 3 $7\%$ \\
\texttt{scaled\_heading\_24room} & $1055$ & $0.24$ & kitchen $10\%$, meeting room 1 $7\%$, meeting room 2 $6\%$ \\
\texttt{agent\_cued\_24room} & $1081$ & $0.19$ & meeting room 1 $8\%$, meeting room 2 $7\%$, kitchen $7\%$ \\
\texttt{agent\_cued\_flat\_24room} & $1107$ & $0.12$ & meeting room 1 $7\%$, lounge $6\%$, meeting room 3 $5\%$ \\
\bottomrule
\end{tabular}
\end{table}

\subsection{Metrics}
\label{app:curiosity-metrics}

Three of the four reported quantities are \emph{conditional on presence}, which we
state explicitly because it determines how the numbers must be read.

\begin{itemize}
  \item \textbf{Recall} --- major-change recall \emph{given presence}:
        $\mathrm{TP}_{\text{major}}/(\mathrm{TP}_{\text{major}}+\mathrm{FN}_{\text{major}})$
        over the major changes the robot was present for. It answers ``when the
        robot was there, did it notice?'', not ``how many changes did it see?''.
  \item \textbf{F1} --- change-detection F1 over the steps the robot actually
        observed, i.e.\ of the per-decision judgement ``has this room changed
        since I last saw it?''. Also conditional on presence, and therefore
        \emph{maximised by camping}: a policy that never leaves the busiest room
        scores very well on it. Every table below therefore prints a move budget
        and an unconditional column next to it.
  \item \textbf{Det.} --- agent-relative detection rate: the fraction of cases in
        which the room differed from the agent's \emph{own} last observation of it
        and the detector said so.
  \item \textbf{Yield} --- change events observed per room move.
  \item \textbf{Moves} --- room moves per 3-day run (of at most $182$).
  \item \textbf{Cov.} (unconditional) --- fraction of the reachable change events
        the strategy witnessed, taking the number of steps at which \emph{any}
        room changes as $100\%$.
  \item \textbf{Missed} (unconditional) --- change events never observed: the room
        was not visited after the change, or was visited without a detection.
\end{itemize}

\subsection{Protocol}

We compare nine strategies across all 15 worlds. To isolate navigation performance, the policy controls every stay-or-move decision while \papername{}'s VLM change detector is shared along with its perceived statistics. Each table cell represents the mean across 10 seeds, where each seed alters policy randomness while keeping the environment fixed.

\subsubsection{Per-World Results}

Each table reports, for one world: major-change recall given presence
(\emph{Recall}), change-detection F1 with its standard deviation over seeds,
changes observed per move (\emph{Yield}), the confusion counts of the
change/no-change decision (\emph{TP}, \emph{FP}, \emph{FN}), the move budget
(\emph{Moves}), the changes observed per run (\emph{Changes}), unconditional
coverage (\emph{Cov.}) and the number of change events never observed
(\emph{Missed}).
Bold marks the best F1, the highest yield, the smallest move budget and the
fewest missed changes in each world.

\begin{table}[H]
\centering
\caption{Navigation-strategy comparison on \texttt{default}. Mean over 10 seeds; $\pm$ is the sample standard deviation of F1 over seeds.}
\label{tab:res-default}
\footnotesize
\setlength{\tabcolsep}{4pt}
\begin{tabular}{l c c c r r r r r r r}
\toprule
Strategy & Recall & F1 & Yield & TP & FP & FN & Moves & Changes & Cov. & Missed \\
\midrule
Random & $89.0\%$ & $67.3\%$\,\tiny$\pm3.6$ & $0.43$ & $64$ & $52$ & $10$ & $182$ & $78$ & $45.9\%$ & $19$ \\
Round-Robin & $93.2\%$ & $81.4\%$\,\tiny$\pm0.8$ & $0.92$ & $74$ & $28$ & $6$ & $91$ & $84$ & $49.7\%$ & $25$ \\
Frequency & $89.3\%$ & $72.8\%$\,\tiny$\pm5.4$ & $0.61$ & $71$ & $42$ & $11$ & $139$ & $85$ & $50.4\%$ & $42$ \\
Staleness & $90.4\%$ & $67.1\%$\,\tiny$\pm2.4$ & $0.44$ & $68$ & $58$ & $9$ & $182$ & $81$ & $47.7\%$ & $18$ \\
Active Mapping & $88.0\%$ & $66.8\%$\,\tiny$\pm2.4$ & $0.44$ & $68$ & $57$ & $10$ & $182$ & $81$ & $47.8\%$ & $\mathbf{15}$ \\
Greedy Hazard & $86.6\%$ & $79.4\%$\,\tiny$\pm1.0$ & $0.65$ & $78$ & $28$ & $12$ & $144$ & $93$ & $55.3\%$ & $76$ \\
Entropy & $85.8\%$ & $91.7\%$\,\tiny$\pm3.4$ & $\mathbf{14.49}$ & $78$ & $1$ & $13$ & $\mathbf{6}$ & $94$ & $55.7\%$ & $383$ \\
Beta-Entropy & $93.3\%$ & $73.3\%$\,\tiny$\pm3.7$ & $0.46$ & $72$ & $44$ & $8$ & $182$ & $84$ & $49.5\%$ & $47$ \\
\textbf{\papername{}} & $97.1\%$ & $\mathbf{92.5\%}$\,\tiny$\pm1.6$ & $2.34$ & $69$ & $9$ & $2$ & $32$ & $74$ & $43.8\%$ & $58$ \\
\bottomrule
\end{tabular}
\end{table}

\begin{table}[H]
\centering
\caption{Navigation-strategy comparison on \texttt{regime}. Mean over 10 seeds; $\pm$ is the sample standard deviation of F1 over seeds.}
\label{tab:res-regime}
\footnotesize
\setlength{\tabcolsep}{4pt}
\begin{tabular}{l c c c r r r r r r r}
\toprule
Strategy & Recall & F1 & Yield & TP & FP & FN & Moves & Changes & Cov. & Missed \\
\midrule
Random & $91.3\%$ & $72.9\%$\,\tiny$\pm2.8$ & $0.24$ & $36$ & $20$ & $6$ & $182$ & $44$ & $33.0\%$ & $6$ \\
Round-Robin & $93.9\%$ & $71.6\%$\,\tiny$\pm1.0$ & $0.40$ & $30$ & $19$ & $4$ & $91$ & $36$ & $26.9\%$ & $\mathbf{5}$ \\
Frequency & $85.5\%$ & $73.1\%$\,\tiny$\pm2.0$ & $0.34$ & $26$ & $12$ & $8$ & $104$ & $35$ & $26.4\%$ & $48$ \\
Staleness & $90.7\%$ & $72.2\%$\,\tiny$\pm3.4$ & $0.22$ & $33$ & $21$ & $5$ & $182$ & $40$ & $29.8\%$ & $7$ \\
Active Mapping & $95.2\%$ & $72.9\%$\,\tiny$\pm3.8$ & $0.22$ & $33$ & $20$ & $5$ & $182$ & $40$ & $29.7\%$ & $7$ \\
Greedy Hazard & $80.4\%$ & $83.0\%$\,\tiny$\pm2.8$ & $0.61$ & $20$ & $2$ & $6$ & $45$ & $28$ & $20.6\%$ & $176$ \\
Entropy & $81.9\%$ & $85.5\%$\,\tiny$\pm1.8$ & $\mathbf{3.71}$ & $25$ & $0$ & $8$ & $\mathbf{10}$ & $35$ & $26.3\%$ & $191$ \\
Beta-Entropy & $90.3\%$ & $74.9\%$\,\tiny$\pm3.7$ & $0.32$ & $44$ & $17$ & $12$ & $182$ & $58$ & $43.6\%$ & $9$ \\
\textbf{\papername{}} & $96.2\%$ & $\mathbf{89.8\%}$\,\tiny$\pm3.7$ & $1.45$ & $40$ & $8$ & $1$ & $30$ & $43$ & $32.3\%$ & $47$ \\
\bottomrule
\end{tabular}
\end{table}

\begin{table}[H]
\centering
\caption{Navigation-strategy comparison on \texttt{adversarial}. Mean over 10 seeds; $\pm$ is the sample standard deviation of F1 over seeds.}
\label{tab:res-adversarial}
\footnotesize
\setlength{\tabcolsep}{4pt}
\begin{tabular}{l c c c r r r r r r r}
\toprule
Strategy & Recall & F1 & Yield & TP & FP & FN & Moves & Changes & Cov. & Missed \\
\midrule
Random & $90.7\%$ & $71.4\%$\,\tiny$\pm3.8$ & $0.33$ & $52$ & $36$ & $5$ & $182$ & $60$ & $47.4\%$ & $19$ \\
Round-Robin & $86.1\%$ & $65.2\%$\,\tiny$\pm0.5$ & $0.57$ & $43$ & $39$ & $8$ & $91$ & $52$ & $40.9\%$ & $21$ \\
Frequency & $88.3\%$ & $77.7\%$\,\tiny$\pm2.8$ & $0.56$ & $63$ & $26$ & $10$ & $135$ & $75$ & $59.2\%$ & $53$ \\
Staleness & $91.6\%$ & $68.9\%$\,\tiny$\pm2.1$ & $0.33$ & $52$ & $42$ & $5$ & $182$ & $59$ & $46.7\%$ & $\mathbf{11}$ \\
Active Mapping & $89.5\%$ & $69.8\%$\,\tiny$\pm1.6$ & $0.33$ & $52$ & $40$ & $5$ & $182$ & $60$ & $47.2\%$ & $14$ \\
Greedy Hazard & $88.5\%$ & $85.4\%$\,\tiny$\pm4.9$ & $0.55$ & $47$ & $9$ & $8$ & $104$ & $57$ & $44.9\%$ & $229$ \\
Entropy & $88.1\%$ & $90.8\%$\,\tiny$\pm2.1$ & $\mathbf{10.45}$ & $62$ & $0$ & $12$ & $\mathbf{7}$ & $76$ & $60.1\%$ & $283$ \\
Beta-Entropy & $88.8\%$ & $78.8\%$\,\tiny$\pm4.6$ & $0.40$ & $61$ & $23$ & $10$ & $182$ & $73$ & $57.5\%$ & $43$ \\
\textbf{\papername{}} & $96.8\%$ & $\mathbf{91.6\%}$\,\tiny$\pm4.1$ & $2.24$ & $50$ & $7$ & $1$ & $24$ & $54$ & $42.4\%$ & $102$ \\
\bottomrule
\end{tabular}
\end{table}

\begin{table}[H]
\centering
\caption{Navigation-strategy comparison on \texttt{base\_v2}. Mean over 10 seeds; $\pm$ is the sample standard deviation of F1 over seeds.}
\label{tab:res-base-v2}
\footnotesize
\setlength{\tabcolsep}{4pt}
\begin{tabular}{l c c c r r r r r r r}
\toprule
Strategy & Recall & F1 & Yield & TP & FP & FN & Moves & Changes & Cov. & Missed \\
\midrule
Random & $87.9\%$ & $63.7\%$\,\tiny$\pm3.4$ & $0.35$ & $51$ & $48$ & $11$ & $182$ & $64$ & $40.3\%$ & $\mathbf{11}$ \\
Round-Robin & $92.5\%$ & $63.1\%$\,\tiny$\pm3.1$ & $0.62$ & $42$ & $37$ & $12$ & $91$ & $57$ & $35.8\%$ & $26$ \\
Frequency & $90.5\%$ & $66.7\%$\,\tiny$\pm4.1$ & $0.50$ & $55$ & $40$ & $14$ & $140$ & $70$ & $44.6\%$ & $43$ \\
Staleness & $88.9\%$ & $64.7\%$\,\tiny$\pm3.6$ & $0.36$ & $58$ & $58$ & $5$ & $182$ & $65$ & $41.1\%$ & $14$ \\
Active Mapping & $88.8\%$ & $66.2\%$\,\tiny$\pm1.8$ & $0.36$ & $59$ & $55$ & $5$ & $182$ & $65$ & $41.2\%$ & $14$ \\
Greedy Hazard & $91.1\%$ & $71.4\%$\,\tiny$\pm3.8$ & $0.65$ & $49$ & $24$ & $15$ & $100$ & $65$ & $41.4\%$ & $121$ \\
Entropy & $79.1\%$ & $74.0\%$\,\tiny$\pm7.4$ & $\mathbf{9.09}$ & $42$ & $2$ & $25$ & $\mathbf{8}$ & $68$ & $43.2\%$ & $319$ \\
Beta-Entropy & $94.1\%$ & $68.8\%$\,\tiny$\pm4.2$ & $0.36$ & $53$ & $37$ & $10$ & $182$ & $65$ & $41.0\%$ & $45$ \\
\textbf{\papername{}} & $99.3\%$ & $\mathbf{81.8\%}$\,\tiny$\pm4.0$ & $2.27$ & $39$ & $8$ & $10$ & $22$ & $51$ & $32.3\%$ & $130$ \\
\bottomrule
\end{tabular}
\end{table}

\begin{table}[H]
\centering
\caption{Navigation-strategy comparison on \texttt{distractor}. Mean over 10 seeds; $\pm$ is the sample standard deviation of F1 over seeds.}
\label{tab:res-distractor}
\footnotesize
\setlength{\tabcolsep}{4pt}
\begin{tabular}{l c c c r r r r r r r}
\toprule
Strategy & Recall & F1 & Yield & TP & FP & FN & Moves & Changes & Cov. & Missed \\
\midrule
Random & $92.8\%$ & $66.7\%$\,\tiny$\pm4.8$ & $0.37$ & $59$ & $52$ & $6$ & $182$ & $68$ & $43.1\%$ & $12$ \\
Round-Robin & $92.8\%$ & $63.3\%$\,\tiny$\pm1.0$ & $0.59$ & $43$ & $40$ & $10$ & $91$ & $54$ & $34.4\%$ & $13$ \\
Frequency & $92.0\%$ & $74.2\%$\,\tiny$\pm4.6$ & $0.69$ & $74$ & $38$ & $14$ & $129$ & $89$ & $56.8\%$ & $45$ \\
Staleness & $94.1\%$ & $60.7\%$\,\tiny$\pm3.3$ & $0.34$ & $56$ & $68$ & $4$ & $182$ & $62$ & $39.4\%$ & $\mathbf{8}$ \\
Active Mapping & $93.2\%$ & $59.2\%$\,\tiny$\pm2.7$ & $0.34$ & $55$ & $69$ & $6$ & $182$ & $62$ & $39.7\%$ & $\mathbf{8}$ \\
Greedy Hazard & $88.8\%$ & $76.8\%$\,\tiny$\pm4.9$ & $0.65$ & $74$ & $30$ & $15$ & $139$ & $90$ & $57.5\%$ & $138$ \\
Entropy & $81.4\%$ & $77.1\%$\,\tiny$\pm13.3$ & $\mathbf{13.25}$ & $58$ & $2$ & $25$ & $\mathbf{6}$ & $85$ & $54.0\%$ & $323$ \\
Beta-Entropy & $93.0\%$ & $68.5\%$\,\tiny$\pm1.8$ & $0.38$ & $57$ & $43$ & $10$ & $182$ & $68$ & $43.6\%$ & $37$ \\
\textbf{\papername{}} & $98.6\%$ & $\mathbf{83.0\%}$\,\tiny$\pm4.4$ & $2.07$ & $44$ & $8$ & $8$ & $26$ & $55$ & $34.8\%$ & $53$ \\
\bottomrule
\end{tabular}
\end{table}

\begin{table}[H]
\centering
\caption{Navigation-strategy comparison on \texttt{intent\_cued}. Mean over 10 seeds; $\pm$ is the sample standard deviation of F1 over seeds.}
\label{tab:res-intent-cued}
\footnotesize
\setlength{\tabcolsep}{4pt}
\begin{tabular}{l c c c r r r r r r r}
\toprule
Strategy & Recall & F1 & Yield & TP & FP & FN & Moves & Changes & Cov. & Missed \\
\midrule
Random & $93.7\%$ & $66.0\%$\,\tiny$\pm4.2$ & $0.34$ & $54$ & $49$ & $6$ & $182$ & $63$ & $39.4\%$ & $12$ \\
Round-Robin & $93.5\%$ & $67.5\%$\,\tiny$\pm3.9$ & $0.65$ & $47$ & $36$ & $9$ & $91$ & $59$ & $37.1\%$ & $15$ \\
Frequency & $89.5\%$ & $70.2\%$\,\tiny$\pm3.8$ & $0.56$ & $59$ & $37$ & $13$ & $132$ & $74$ & $46.7\%$ & $44$ \\
Staleness & $90.1\%$ & $62.5\%$\,\tiny$\pm3.5$ & $0.34$ & $54$ & $59$ & $6$ & $182$ & $63$ & $39.4\%$ & $10$ \\
Active Mapping & $90.3\%$ & $61.8\%$\,\tiny$\pm1.4$ & $0.34$ & $54$ & $61$ & $6$ & $182$ & $62$ & $39.2\%$ & $\mathbf{9}$ \\
Greedy Hazard & $88.1\%$ & $73.0\%$\,\tiny$\pm3.6$ & $0.68$ & $45$ & $21$ & $12$ & $87$ & $59$ & $37.4\%$ & $74$ \\
Entropy & $88.2\%$ & $\mathbf{82.8\%}$\,\tiny$\pm4.8$ & $\mathbf{11.58}$ & $56$ & $1$ & $22$ & $\mathbf{7}$ & $80$ & $50.9\%$ & $308$ \\
Beta-Entropy & $92.1\%$ & $68.3\%$\,\tiny$\pm3.6$ & $0.34$ & $52$ & $40$ & $9$ & $182$ & $63$ & $39.4\%$ & $30$ \\
\textbf{\papername{}} & $98.3\%$ & $80.3\%$\,\tiny$\pm6.2$ & $1.92$ & $38$ & $10$ & $8$ & $25$ & $49$ & $30.7\%$ & $87$ \\
\bottomrule
\end{tabular}
\end{table}

\begin{table}[H]
\centering
\caption{Navigation-strategy comparison on \texttt{intent\_cued\_heading}. Mean over 10 seeds; $\pm$ is the sample standard deviation of F1 over seeds.}
\label{tab:res-intent-cued-heading}
\footnotesize
\setlength{\tabcolsep}{4pt}
\begin{tabular}{l c c c r r r r r r r}
\toprule
Strategy & Recall & F1 & Yield & TP & FP & FN & Moves & Changes & Cov. & Missed \\
\midrule
Random & $91.0\%$ & $66.3\%$\,\tiny$\pm4.5$ & $0.37$ & $56$ & $47$ & $9$ & $182$ & $67$ & $41.9\%$ & $14$ \\
Round-Robin & $94.1\%$ & $65.8\%$\,\tiny$\pm2.3$ & $0.59$ & $44$ & $39$ & $8$ & $91$ & $54$ & $34.0\%$ & $34$ \\
Frequency & $91.9\%$ & $72.0\%$\,\tiny$\pm3.6$ & $0.58$ & $64$ & $36$ & $13$ & $135$ & $79$ & $49.6\%$ & $44$ \\
Staleness & $85.4\%$ & $63.1\%$\,\tiny$\pm2.9$ & $0.36$ & $54$ & $53$ & $10$ & $182$ & $66$ & $41.8\%$ & $11$ \\
Active Mapping & $86.4\%$ & $60.9\%$\,\tiny$\pm2.6$ & $0.37$ & $55$ & $60$ & $10$ & $182$ & $67$ & $42.3\%$ & $\mathbf{10}$ \\
Greedy Hazard & $90.1\%$ & $77.9\%$\,\tiny$\pm2.2$ & $0.67$ & $66$ & $25$ & $13$ & $120$ & $80$ & $50.6\%$ & $121$ \\
Entropy & $85.1\%$ & $82.1\%$\,\tiny$\pm3.4$ & $\mathbf{14.10}$ & $60$ & $1$ & $24$ & $\mathbf{6}$ & $86$ & $54.1\%$ & $319$ \\
Beta-Entropy & $85.3\%$ & $67.7\%$\,\tiny$\pm3.7$ & $0.39$ & $55$ & $40$ & $13$ & $182$ & $70$ & $44.1\%$ & $51$ \\
\textbf{\papername{}} & $97.9\%$ & $\mathbf{82.7\%}$\,\tiny$\pm4.6$ & $2.36$ & $44$ & $8$ & $10$ & $24$ & $56$ & $35.5\%$ & $95$ \\
\bottomrule
\end{tabular}
\end{table}

\begin{table}[H]
\centering
\caption{Navigation-strategy comparison on \texttt{agent\_cued\_6room}. Mean over 10 seeds; $\pm$ is the sample standard deviation of F1 over seeds.}
\label{tab:res-agent-cued-6room}
\footnotesize
\setlength{\tabcolsep}{4pt}
\begin{tabular}{l c c c r r r r r r r}
\toprule
Strategy & Recall & F1 & Yield & TP & FP & FN & Moves & Changes & Cov. & Missed \\
\midrule
Random & $95.3\%$ & $66.7\%$\,\tiny$\pm3.4$ & $0.35$ & $55$ & $49$ & $6$ & $182$ & $64$ & $41.8\%$ & $15$ \\
Round-Robin & $97.8\%$ & $68.1\%$\,\tiny$\pm0.8$ & $0.64$ & $47$ & $35$ & $9$ & $91$ & $58$ & $38.2\%$ & $25$ \\
Frequency & $94.8\%$ & $69.9\%$\,\tiny$\pm3.8$ & $0.51$ & $58$ & $41$ & $9$ & $135$ & $69$ & $45.5\%$ & $35$ \\
Staleness & $91.1\%$ & $62.8\%$\,\tiny$\pm3.7$ & $0.36$ & $57$ & $60$ & $6$ & $182$ & $65$ & $42.7\%$ & $\mathbf{6}$ \\
Active Mapping & $92.3\%$ & $65.2\%$\,\tiny$\pm2.8$ & $0.37$ & $59$ & $57$ & $6$ & $182$ & $67$ & $44.0\%$ & $7$ \\
Greedy Hazard & $91.2\%$ & $71.9\%$\,\tiny$\pm4.1$ & $0.61$ & $42$ & $24$ & $9$ & $88$ & $54$ & $35.5\%$ & $118$ \\
Entropy & $83.2\%$ & $78.4\%$\,\tiny$\pm5.0$ & $\mathbf{9.58}$ & $41$ & $2$ & $19$ & $\mathbf{6}$ & $62$ & $41.0\%$ & $314$ \\
Beta-Entropy & $94.2\%$ & $67.8\%$\,\tiny$\pm4.4$ & $0.34$ & $51$ & $40$ & $9$ & $182$ & $62$ & $40.7\%$ & $35$ \\
\textbf{\papername{}} & $99.2\%$ & $\mathbf{80.2\%}$\,\tiny$\pm6.7$ & $2.20$ & $39$ & $9$ & $10$ & $23$ & $52$ & $33.9\%$ & $113$ \\
\bottomrule
\end{tabular}
\end{table}

\begin{table}[H]
\centering
\caption{Navigation-strategy comparison on \texttt{agent\_cued\_flat\_6room}. Mean over 10 seeds; $\pm$ is the sample standard deviation of F1 over seeds.}
\label{tab:res-agent-cued-flat-6room}
\footnotesize
\setlength{\tabcolsep}{4pt}
\begin{tabular}{l c c c r r r r r r r}
\toprule
Strategy & Recall & F1 & Yield & TP & FP & FN & Moves & Changes & Cov. & Missed \\
\midrule
Random & $89.1\%$ & $65.1\%$\,\tiny$\pm3.9$ & $0.39$ & $59$ & $53$ & $9$ & $182$ & $70$ & $45.1\%$ & $10$ \\
Round-Robin & $99.3\%$ & $63.7\%$\,\tiny$\pm0.5$ & $0.52$ & $41$ & $41$ & $6$ & $91$ & $47$ & $30.1\%$ & $16$ \\
Frequency & $91.8\%$ & $68.3\%$\,\tiny$\pm4.8$ & $0.46$ & $54$ & $41$ & $9$ & $141$ & $66$ & $42.0\%$ & $35$ \\
Staleness & $90.5\%$ & $59.2\%$\,\tiny$\pm2.7$ & $0.35$ & $56$ & $71$ & $6$ & $182$ & $64$ & $40.9\%$ & $6$ \\
Active Mapping & $86.8\%$ & $57.5\%$\,\tiny$\pm2.2$ & $0.34$ & $52$ & $70$ & $7$ & $182$ & $62$ & $39.5\%$ & $\mathbf{5}$ \\
Greedy Hazard & $85.9\%$ & $69.1\%$\,\tiny$\pm2.6$ & $0.62$ & $39$ & $26$ & $9$ & $82$ & $50$ & $32.3\%$ & $135$ \\
Entropy & $85.6\%$ & $79.8\%$\,\tiny$\pm5.0$ & $\mathbf{10.82}$ & $44$ & $2$ & $21$ & $\mathbf{6}$ & $67$ & $43.0\%$ & $335$ \\
Beta-Entropy & $92.0\%$ & $65.3\%$\,\tiny$\pm2.9$ & $0.36$ & $53$ & $45$ & $11$ & $182$ & $66$ & $42.0\%$ & $36$ \\
\textbf{\papername{}} & $99.2\%$ & $\mathbf{81.1\%}$\,\tiny$\pm4.9$ & $2.39$ & $51$ & $11$ & $12$ & $28$ & $66$ & $42.4\%$ & $102$ \\
\bottomrule
\end{tabular}
\end{table}

\begin{table}[H]
\centering
\caption{Navigation-strategy comparison on \texttt{scaled\_base\_24room}. Mean over 10 seeds; $\pm$ is the sample standard deviation of F1 over seeds.}
\label{tab:res-scaled-base-24room}
\footnotesize
\setlength{\tabcolsep}{4pt}
\begin{tabular}{l c c c r r r r r r r}
\toprule
Strategy & Recall & F1 & Yield & TP & FP & FN & Moves & Changes & Cov. & Missed \\
\midrule
Random & $90.4\%$ & $62.1\%$\,\tiny$\pm3.6$ & $0.38$ & $55$ & $62$ & $5$ & $182$ & $68$ & $37.7\%$ & $227$ \\
Round-Robin & $95.1\%$ & $71.4\%$\,\tiny$\pm1.4$ & $0.77$ & $50$ & $27$ & $12$ & $91$ & $70$ & $38.7\%$ & $211$ \\
Frequency & $92.7\%$ & $67.8\%$\,\tiny$\pm6.0$ & $0.45$ & $61$ & $51$ & $6$ & $168$ & $75$ & $41.5\%$ & $584$ \\
Staleness & $89.2\%$ & $61.8\%$\,\tiny$\pm2.9$ & $0.41$ & $58$ & $67$ & $6$ & $182$ & $74$ & $40.8\%$ & $148$ \\
Active Mapping & $91.7\%$ & $60.1\%$\,\tiny$\pm3.0$ & $0.38$ & $56$ & $70$ & $4$ & $182$ & $69$ & $38.3\%$ & $\mathbf{142}$ \\
Greedy Hazard & $94.9\%$ & $80.8\%$\,\tiny$\pm6.7$ & $0.68$ & $67$ & $20$ & $8$ & $122$ & $83$ & $45.6\%$ & $1422$ \\
Entropy & $82.8\%$ & $78.0\%$\,\tiny$\pm7.2$ & $\mathbf{3.18}$ & $48$ & $5$ & $20$ & $\mathbf{25}$ & $78$ & $43.2\%$ & $1565$ \\
Beta-Entropy & $91.3\%$ & $75.1\%$\,\tiny$\pm3.7$ & $0.40$ & $52$ & $24$ & $11$ & $182$ & $73$ & $40.2\%$ & $965$ \\
\textbf{\papername{}} & $99.5\%$ & $\mathbf{84.2\%}$\,\tiny$\pm5.1$ & $2.32$ & $50$ & $6$ & $13$ & $30$ & $70$ & $39.0\%$ & $1033$ \\
\bottomrule
\end{tabular}
\end{table}

\begin{table}[H]
\centering
\caption{Navigation-strategy comparison on \texttt{scaled\_24room}. Mean over 10 seeds; $\pm$ is the sample standard deviation of F1 over seeds.}
\label{tab:res-scaled-24room}
\footnotesize
\setlength{\tabcolsep}{4pt}
\begin{tabular}{l c c c r r r r r r r}
\toprule
Strategy & Recall & F1 & Yield & TP & FP & FN & Moves & Changes & Cov. & Missed \\
\midrule
Random & $92.1\%$ & $61.5\%$\,\tiny$\pm3.3$ & $0.37$ & $54$ & $62$ & $5$ & $182$ & $68$ & $37.5\%$ & $269$ \\
Round-Robin & $92.8\%$ & $67.1\%$\,\tiny$\pm1.7$ & $0.57$ & $37$ & $29$ & $7$ & $91$ & $52$ & $28.7\%$ & $327$ \\
Frequency & $95.9\%$ & $69.1\%$\,\tiny$\pm4.9$ & $0.46$ & $63$ & $50$ & $5$ & $167$ & $77$ & $42.3\%$ & $568$ \\
Staleness & $92.4\%$ & $61.5\%$\,\tiny$\pm3.8$ & $0.38$ & $57$ & $68$ & $4$ & $182$ & $69$ & $38.0\%$ & $\mathbf{150}$ \\
Active Mapping & $93.6\%$ & $62.5\%$\,\tiny$\pm3.1$ & $0.38$ & $59$ & $67$ & $4$ & $182$ & $70$ & $38.6\%$ & $157$ \\
Greedy Hazard & $88.7\%$ & $76.7\%$\,\tiny$\pm9.7$ & $0.64$ & $57$ & $20$ & $10$ & $116$ & $74$ & $40.6\%$ & $1277$ \\
Entropy & $79.4\%$ & $75.4\%$\,\tiny$\pm4.6$ & $\mathbf{2.90}$ & $42$ & $5$ & $22$ & $\mathbf{24}$ & $71$ & $39.1\%$ & $1559$ \\
Beta-Entropy & $91.3\%$ & $75.2\%$\,\tiny$\pm3.2$ & $0.38$ & $52$ & $22$ & $12$ & $182$ & $70$ & $38.5\%$ & $936$ \\
\textbf{\papername{}} & $99.4\%$ & $\mathbf{84.4\%}$\,\tiny$\pm4.9$ & $1.98$ & $54$ & $9$ & $12$ & $38$ & $75$ & $41.6\%$ & $730$ \\
\bottomrule
\end{tabular}
\end{table}

\begin{table}[H]
\centering
\caption{Navigation-strategy comparison on \texttt{scaled\_distractor\_24room}. Mean over 10 seeds; $\pm$ is the sample standard deviation of F1 over seeds.}
\label{tab:res-scaled-distractor-24room}
\footnotesize
\setlength{\tabcolsep}{4pt}
\begin{tabular}{l c c c r r r r r r r}
\toprule
Strategy & Recall & F1 & Yield & TP & FP & FN & Moves & Changes & Cov. & Missed \\
\midrule
Random & $91.5\%$ & $62.2\%$\,\tiny$\pm3.1$ & $0.39$ & $56$ & $63$ & $6$ & $182$ & $71$ & $39.0\%$ & $263$ \\
Round-Robin & $95.7\%$ & $67.6\%$\,\tiny$\pm1.2$ & $0.74$ & $50$ & $38$ & $10$ & $91$ & $67$ & $37.1\%$ & $252$ \\
Frequency & $96.2\%$ & $72.3\%$\,\tiny$\pm6.3$ & $0.49$ & $70$ & $48$ & $4$ & $167$ & $82$ & $45.2\%$ & $594$ \\
Staleness & $92.3\%$ & $60.9\%$\,\tiny$\pm3.8$ & $0.39$ & $58$ & $70$ & $4$ & $182$ & $71$ & $39.3\%$ & $154$ \\
Active Mapping & $95.1\%$ & $63.2\%$\,\tiny$\pm3.8$ & $0.40$ & $60$ & $66$ & $4$ & $182$ & $72$ & $40.0\%$ & $\mathbf{152}$ \\
Greedy Hazard & $94.1\%$ & $84.4\%$\,\tiny$\pm6.5$ & $0.69$ & $88$ & $22$ & $8$ & $152$ & $105$ & $57.8\%$ & $1353$ \\
Entropy & $75.4\%$ & $72.9\%$\,\tiny$\pm7.3$ & $\mathbf{2.79}$ & $38$ & $6$ & $22$ & $\mathbf{25}$ & $69$ & $38.2\%$ & $1615$ \\
Beta-Entropy & $94.5\%$ & $76.8\%$\,\tiny$\pm4.0$ & $0.39$ & $54$ & $23$ & $10$ & $182$ & $71$ & $39.1\%$ & $979$ \\
\textbf{\papername{}} & $99.8\%$ & $\mathbf{87.6\%}$\,\tiny$\pm6.7$ & $2.21$ & $56$ & $7$ & $10$ & $34$ & $74$ & $41.2\%$ & $937$ \\
\bottomrule
\end{tabular}
\end{table}

\begin{table}[H]
\centering
\caption{Navigation-strategy comparison on \texttt{scaled\_heading\_24room}. Mean over 10 seeds; $\pm$ is the sample standard deviation of F1 over seeds.}
\label{tab:res-scaled-heading-24room}
\footnotesize
\setlength{\tabcolsep}{4pt}
\begin{tabular}{l c c c r r r r r r r}
\toprule
Strategy & Recall & F1 & Yield & TP & FP & FN & Moves & Changes & Cov. & Missed \\
\midrule
Random & $91.0\%$ & $60.3\%$\,\tiny$\pm4.4$ & $0.36$ & $52$ & $63$ & $5$ & $182$ & $65$ & $36.0\%$ & $246$ \\
Round-Robin & $95.2\%$ & $65.3\%$\,\tiny$\pm2.5$ & $0.69$ & $43$ & $36$ & $9$ & $91$ & $63$ & $34.8\%$ & $270$ \\
Frequency & $91.5\%$ & $66.5\%$\,\tiny$\pm4.5$ & $0.45$ & $61$ & $53$ & $8$ & $169$ & $76$ & $41.9\%$ & $535$ \\
Staleness & $93.3\%$ & $60.9\%$\,\tiny$\pm3.8$ & $0.38$ & $56$ & $67$ & $5$ & $182$ & $69$ & $38.1\%$ & $158$ \\
Active Mapping & $92.3\%$ & $60.0\%$\,\tiny$\pm4.9$ & $0.36$ & $54$ & $67$ & $5$ & $182$ & $66$ & $36.4\%$ & $\mathbf{153}$ \\
Greedy Hazard & $93.0\%$ & $75.5\%$\,\tiny$\pm4.9$ & $0.63$ & $49$ & $22$ & $9$ & $104$ & $65$ & $36.1\%$ & $1442$ \\
Entropy & $81.2\%$ & $78.7\%$\,\tiny$\pm4.9$ & $\mathbf{2.78}$ & $43$ & $6$ & $17$ & $\mathbf{25}$ & $68$ & $37.7\%$ & $1559$ \\
Beta-Entropy & $93.6\%$ & $74.0\%$\,\tiny$\pm3.6$ & $0.37$ & $48$ & $23$ & $10$ & $182$ & $66$ & $36.7\%$ & $1052$ \\
\textbf{\papername{}} & $99.2\%$ & $\mathbf{85.9\%}$\,\tiny$\pm2.9$ & $1.96$ & $50$ & $7$ & $10$ & $36$ & $70$ & $38.7\%$ & $819$ \\
\bottomrule
\end{tabular}
\end{table}

\begin{table}[H]
\centering
\caption{Navigation-strategy comparison on \texttt{agent\_cued\_24room}. Mean over 10 seeds; $\pm$ is the sample standard deviation of F1 over seeds.}
\label{tab:res-agent-cued-24room}
\footnotesize
\setlength{\tabcolsep}{4pt}
\begin{tabular}{l c c c r r r r r r r}
\toprule
Strategy & Recall & F1 & Yield & TP & FP & FN & Moves & Changes & Cov. & Missed \\
\midrule
Random & $88.2\%$ & $61.1\%$\,\tiny$\pm5.3$ & $0.37$ & $54$ & $63$ & $6$ & $182$ & $67$ & $37.2\%$ & $244$ \\
Round-Robin & $89.9\%$ & $67.1\%$\,\tiny$\pm3.7$ & $0.76$ & $46$ & $32$ & $13$ & $91$ & $69$ & $38.1\%$ & $255$ \\
Frequency & $90.4\%$ & $66.5\%$\,\tiny$\pm4.7$ & $0.44$ & $59$ & $52$ & $7$ & $168$ & $74$ & $40.7\%$ & $583$ \\
Staleness & $90.4\%$ & $61.3\%$\,\tiny$\pm2.4$ & $0.38$ & $57$ & $66$ & $5$ & $182$ & $69$ & $38.2\%$ & $152$ \\
Active Mapping & $92.3\%$ & $61.2\%$\,\tiny$\pm3.3$ & $0.37$ & $55$ & $65$ & $5$ & $182$ & $67$ & $37.0\%$ & $\mathbf{152}$ \\
Greedy Hazard & $91.5\%$ & $76.8\%$\,\tiny$\pm4.1$ & $0.61$ & $62$ & $26$ & $11$ & $131$ & $80$ & $44.5\%$ & $1236$ \\
Entropy & $82.7\%$ & $79.2\%$\,\tiny$\pm5.7$ & $\mathbf{3.11}$ & $50$ & $6$ & $19$ & $\mathbf{24}$ & $76$ & $42.2\%$ & $1574$ \\
Beta-Entropy & $93.1\%$ & $74.1\%$\,\tiny$\pm1.8$ & $0.40$ & $51$ & $24$ & $12$ & $182$ & $72$ & $39.7\%$ & $972$ \\
\textbf{\papername{}} & $99.8\%$ & $\mathbf{84.1\%}$\,\tiny$\pm4.0$ & $1.90$ & $57$ & $10$ & $12$ & $41$ & $77$ & $42.7\%$ & $713$ \\
\bottomrule
\end{tabular}
\end{table}

\begin{table}[H]
\centering
\caption{Navigation-strategy comparison on \texttt{agent\_cued\_flat\_24room}. Mean over 10 seeds; $\pm$ is the sample standard deviation of F1 over seeds.}
\label{tab:res-agent-cued-flat-24room}
\footnotesize
\setlength{\tabcolsep}{4pt}
\begin{tabular}{l c c c r r r r r r r}
\toprule
Strategy & Recall & F1 & Yield & TP & FP & FN & Moves & Changes & Cov. & Missed \\
\midrule
Random & $92.5\%$ & $61.6\%$\,\tiny$\pm3.9$ & $0.38$ & $55$ & $64$ & $5$ & $182$ & $68$ & $37.5\%$ & $272$ \\
Round-Robin & $90.5\%$ & $69.5\%$\,\tiny$\pm1.3$ & $0.76$ & $50$ & $34$ & $10$ & $91$ & $69$ & $37.9\%$ & $221$ \\
Frequency & $91.2\%$ & $63.7\%$\,\tiny$\pm3.9$ & $0.42$ & $56$ & $56$ & $7$ & $169$ & $70$ & $38.7\%$ & $586$ \\
Staleness & $86.0\%$ & $58.4\%$\,\tiny$\pm4.2$ & $0.39$ & $56$ & $73$ & $6$ & $182$ & $70$ & $38.7\%$ & $\mathbf{129}$ \\
Active Mapping & $88.3\%$ & $60.0\%$\,\tiny$\pm4.1$ & $0.38$ & $57$ & $70$ & $6$ & $182$ & $70$ & $38.5\%$ & $139$ \\
Greedy Hazard & $92.5\%$ & $74.5\%$\,\tiny$\pm3.7$ & $0.65$ & $47$ & $21$ & $11$ & $98$ & $63$ & $34.8\%$ & $1489$ \\
Entropy & $77.9\%$ & $75.8\%$\,\tiny$\pm7.6$ & $\mathbf{2.71}$ & $39$ & $6$ & $19$ & $\mathbf{24}$ & $66$ & $36.4\%$ & $1626$ \\
Beta-Entropy & $90.4\%$ & $71.0\%$\,\tiny$\pm4.1$ & $0.39$ & $50$ & $28$ & $12$ & $182$ & $72$ & $39.3\%$ & $1032$ \\
\textbf{\papername{}} & $99.8\%$ & $\mathbf{86.4\%}$\,\tiny$\pm4.1$ & $2.28$ & $55$ & $6$ & $12$ & $32$ & $74$ & $40.7\%$ & $982$ \\
\bottomrule
\end{tabular}
\end{table}

\subsection{Ablations}

Table~\ref{tab:ablations-appendix} removes the navigator, while the
destination is taken as the argmax of the code-computed room-score table instead
of being chosen by the VLM. Also it shows the results if one of the three signals are removed from the prompt.

\begin{table}[H]
\centering
\caption{Ablations of \papername{}, per world: change-detection F1 (\%) with the move budget in
parentheses, mean over 10 seeds. All variants share the same detector and the same dwell policy.
Mean F1 \emph{rises} when a signal is removed, because F1 is conditional on presence and the
ablated variants collapse into camping: at $7.7$ moves the \emph{all three} variant enters $22\%$
of rooms and never observes $812$ changes per run, against $30$ moves, $96\%$ of rooms and $400$
missed for the full policy. The signals therefore buy worst-case robustness and coverage, not mean
F1: the worst case is \texttt{adversarial}, where removing staleness drops F1 from $91.6\%$ at $24$
moves to $62.3\%$ at $6$ moves.}
\label{tab:ablations-appendix}
\footnotesize
\setlength{\tabcolsep}{3pt}
\resizebox{\textwidth}{!}{%
\begin{tabular}{lrrrrrr}
\toprule
World & \textbf{\papername{} (full)} & Score argmax & $-$ hotspots & $-$ expected change & $-$ staleness & $-$ all three \\
\midrule
\texttt{default} & $92.5$ \tiny($32$) & $87.5$ \tiny($81$) & $91.7$ \tiny($42$) & $93.6$ \tiny($26$) & $93.9$ \tiny($6$) & $95.2$ \tiny($3$) \\
\texttt{regime} & $89.8$ \tiny($30$) & $89.8$ \tiny($93$) & $90.8$ \tiny($44$) & $89.6$ \tiny($24$) & $97.5$ \tiny($8$) & $97.1$ \tiny($4$) \\
\texttt{adversarial} & $91.6$ \tiny($24$) & $83.0$ \tiny($109$) & $90.8$ \tiny($35$) & $88.4$ \tiny($20$) & $62.3$ \tiny($6$) & $61.3$ \tiny($4$) \\
\midrule
\texttt{base\_v2} & $81.8$ \tiny($22$) & $71.2$ \tiny($110$) & $77.6$ \tiny($42$) & $82.5$ \tiny($19$) & $85.3$ \tiny($6$) & $85.7$ \tiny($3$) \\
\texttt{distractor} & $83.0$ \tiny($26$) & $74.0$ \tiny($102$) & $80.3$ \tiny($46$) & $78.7$ \tiny($25$) & $88.8$ \tiny($4$) & $90.4$ \tiny($2$) \\
\texttt{intent\_cued} & $80.3$ \tiny($25$) & $73.9$ \tiny($108$) & $80.6$ \tiny($44$) & $80.8$ \tiny($29$) & $89.3$ \tiny($6$) & $90.0$ \tiny($2$) \\
\texttt{intent\_cued\_heading} & $82.7$ \tiny($24$) & $75.3$ \tiny($107$) & $82.0$ \tiny($40$) & $83.8$ \tiny($25$) & $85.2$ \tiny($6$) & $86.4$ \tiny($4$) \\
\texttt{agent\_cued\_6room} & $80.2$ \tiny($23$) & $74.4$ \tiny($109$) & $78.5$ \tiny($43$) & $81.7$ \tiny($28$) & $85.1$ \tiny($7$) & $86.5$ \tiny($5$) \\
\texttt{agent\_cued\_flat\_6room} & $81.1$ \tiny($28$) & $71.4$ \tiny($104$) & $77.8$ \tiny($44$) & $79.4$ \tiny($31$) & $87.5$ \tiny($9$) & $88.0$ \tiny($2$) \\
\midrule
\texttt{scaled\_base\_24room} & $84.2$ \tiny($30$) & $71.9$ \tiny($111$) & $82.3$ \tiny($49$) & $85.9$ \tiny($22$) & $87.2$ \tiny($20$) & $85.9$ \tiny($16$) \\
\texttt{scaled\_24room} & $84.4$ \tiny($38$) & $74.6$ \tiny($111$) & $83.0$ \tiny($46$) & $86.0$ \tiny($23$) & $85.5$ \tiny($17$) & $84.8$ \tiny($16$) \\
\texttt{scaled\_distractor\_24room} & $87.6$ \tiny($34$) & $80.5$ \tiny($92$) & $86.3$ \tiny($45$) & $88.2$ \tiny($18$) & $85.4$ \tiny($22$) & $85.3$ \tiny($17$) \\
\texttt{scaled\_heading\_24room} & $85.9$ \tiny($36$) & $72.2$ \tiny($111$) & $84.4$ \tiny($39$) & $87.4$ \tiny($19$) & $86.6$ \tiny($15$) & $85.9$ \tiny($12$) \\
\texttt{agent\_cued\_24room} & $84.1$ \tiny($41$) & $72.6$ \tiny($112$) & $82.3$ \tiny($47$) & $86.7$ \tiny($24$) & $86.3$ \tiny($18$) & $85.3$ \tiny($14$) \\
\texttt{agent\_cued\_flat\_24room} & $86.4$ \tiny($32$) & $74.0$ \tiny($110$) & $82.6$ \tiny($53$) & $86.4$ \tiny($20$) & $86.5$ \tiny($18$) & $86.7$ \tiny($14$) \\
\midrule
\textit{Mean F1} & $85.0$ & $76.4$ & $83.4$ & $85.3$ & $86.2$ & $86.3$ \\
\textit{Mean moves} & $30$ & $105$ & $44$ & $24$ & $11$ & $8$ \\
\textit{Rooms entered} & $0.96$ & $1.00$ & $0.98$ & $0.86$ & $0.27$ & $0.22$ \\
\textit{Changes missed} & $400$ & $107$ & $302$ & $542$ & $798$ & $812$ \\
\bottomrule
\end{tabular}
}
\end{table}

\subsection{Isolating the destination choice}
In Table~\ref{tab:nav_ablation} each baseline uses its own detector prompt and its own rule for when to move, so the comparison is between complete exploration systems. To isolate the destination choice we re-ran every baseline with \papername{}'s detector and \papername{}'s stay-or-move decision, varying only the rule that selects the next room.

Granting this decision improves every baseline, by 0.120 F1 on average (range 0.056–0.162). The effect is not simply "move less": it releases Entropy from its statistical camp (13 → 45 moves) while restraining the sweeping policies (182 → 74 moves) and the hazard policy (112 → 8). 

Under this condition \papername{}'s destination rule is no longer the strongest on F1 (Table X): a hazard-greedy destination reaches 90.7\% F1 at 8 moves, and four rules clear 80\% on both precision and recall. Those gains come either from camping — the hazard-greedy variant leaves 715 changes unobserved, and Entropy 668 — or from moving two to four times more. COOL leaves the fewest changes unobserved among strategies with a comparable movement budget (389 at 28 moves).
We therefore attribute most of the benefit to the detector's stay-or-move decision, which is transferable, and the remainder to the navigator's additional restraint on movement: argmax retains the same stay decision yet still makes 102 moves against COOL's 28, because it always relocates once the detector releases the robot, whereas the navigator frequently elects to remain.

\begin{table}[t]
\centering
\begin{tabular}{lccccc}
\toprule
Destination rule & Recall & Prec. & F1 & Moves & Missed \\
\midrule
Greedy Hazard           & $84.9\%$ & $\mathbf{98.2\%}$ & $\mathbf{90.7\%}$ & $\mathbf{8}$  & $715$ \\
\textbf{COOL navigator} & $87.6\%$ & $87.0\%$ & $86.8\%$ & $28$    & $\mathbf{389}$ \\
Entropy                 & $87.0\%$ & $87.1\%$ & $86.5\%$ & $45$    & $668$ \\
Frequency               & $90.5\%$ & $81.8\%$ & $85.7\%$ & $67$    & $408$ \\
Beta-Entropy (cost)     & $89.9\%$ & $76.8\%$ & $82.7\%$ & $125$   & $358$ \\
Beta-Entropy            & $88.9\%$ & $75.2\%$ & $81.3\%$ & $61$    & $285$ \\
Round-Robin             & $89.6\%$ & $73.8\%$ & $80.8\%$ & $70$    & $124$ \\
Staleness               & $90.9\%$ & $72.6\%$ & $80.6\%$ & $74$    & $127$ \\
Active Mapping          & $90.9\%$ & $71.9\%$ & $80.2\%$ & $75$    & $122$ \\
Random                  & $91.0\%$ & $70.9\%$ & $79.5\%$ & $104$   & $163$ \\
Score argmax            & $\mathbf{91.6\%}$ & $68.6\%$ & $78.3\%$ & $102$ & $95$ \\
\bottomrule
\end{tabular}
\end{table}

\FloatBarrier
\section{Perception and Re-Identification Ablations}
\label{app:perception_ablations}

\subsection{Object Detection}
\label{app:object_detection_ablation}

We compare several object detection models to identify the most suitable model for low-quality images captured by the robot. The evaluation is conducted on a self-labeled dataset collected in the office environment in which the robot is deployed. For each model, we report the best-performing confidence threshold based on multiple test runs.

Since the downstream task requires reliable detections, precision is prioritized over recall. In this setting, false positives are more problematic than false negatives, because incorrect detections may lead to erroneous robot behavior. Therefore, we use the F$_{0.5}$ score as the main selection criterion, as it gives more weight to precision than recall. Table~\ref{tab:object-detection-performance} shows the results.
\begin{table}[h]
\centering
\renewcommand{\arraystretch}{1.1}
\caption{Object detection performance on the self-labeled office dataset. The confidence threshold is selected based on the best performance observed across multiple test runs.}
\label{tab:object-detection-performance}
\resizebox{0.7\textwidth}{!}{%
\begin{tabular}{lrrrrrrr}
\hline
Model & Conf & TP & FP & FN & Precision & Recall & F$_{0.5}$ \\
\hline
GroundingDINO \citep{liu2023grounding} & 0.55 & 11.0 & 2.0 & 147.0 & 0.846 & 0.070 & 0.262 \\
YOLO11m-seg & 0.55 & 88.0 & 2.3 & 70.0 & 0.975 & 0.557 & 0.848 \\
YOLO11l-seg & 0.55 & 88.0 & 3.0 & 70.0 & 0.967 & 0.557 & 0.843 \\
YOLO11s-seg & 0.40 & 87.0 & 3.0 & 71.0 & 0.967 & 0.551 & 0.840 \\
OWLv2 \citep{minderer2024scalingopenvocabularyobjectdetection} & 0.40 & 19.3 & 0.3 & 138.7 & 0.974 & 0.122 & 0.402 \\
\hline
\end{tabular}%
}
\end{table}

\subsection{Person Re-Identification}
\label{app:reid_ablation}
We evaluate several feature extractors for person re-identification on a self-labeled test set containing 20 identities. For each model and input variant, the threshold with the best performance across multiple test runs is reported. The input variants include the original image, Lanczos upscaling~\citep{LanczosFilteringinOneandTwoDimensions}, Real-ESRGAN upscaling~\citep{wang2021realesrgan}, and a preprocessing variant using white balancing and brightness adjustment.

The results are shown in Table~\ref{tab:full_clustering_ablation}. Overall, the fine-tuned OSNet model achieves the best clustering performance. In particular, using the original whole-image input yields the highest Pairwise F1 score, purity, and precision, while Real-ESRGAN upscaling produces the highest recall.
\begin{table}[H]
\centering
\scriptsize
\setlength{\tabcolsep}{4pt}
\renewcommand{\arraystretch}{1.1}
\caption{Person re-identification clustering ablation results on a test set with 20 identities. For each model and input variant, the best-performing threshold across multiple test runs is reported. Bold values indicate the best result per metric, and underlined values indicate the second-best result.}
\label{tab:full_clustering_ablation}
\begin{tabular}{llrrrrrr}
\toprule
Model & Input & Threshold & Clusters & Pairwise F1 & Purity & Precision & Recall \\
\midrule
ConvNeXt-Tiny \citep{convnext} & Whole Image & 0.500 & 60 & 0.151 & 0.500 & 0.159 & 0.144 \\
ConvNeXt-Tiny \citep{convnext}& Lanczos Upscaled & 0.500 & 60 & 0.147 & 0.496 & 0.154 & 0.141 \\
ConvNeXt-Tiny \citep{convnext}& Whole Image + WB + Bright & 0.500 & 85 & 0.122 & 0.572 & 0.166 & 0.097 \\
ConvNeXt-Tiny \citep{convnext}& Real-ESRGAN Upscaled & 0.500 & 93 & 0.105 & 0.572 & 0.129 & 0.089 \\
\midrule
DINOv3 \citep{dinov3} & Lanczos Upscaled & 0.400 & 6 & 0.180 & 0.228 & 0.099 & 0.925 \\
DINOv3 \citep{dinov3} & Real-ESRGAN Upscaled & 0.450 & 6 & 0.177 & 0.228 & 0.098 & 0.893 \\
DINOv3 \citep{dinov3} & Whole Image + WB + Bright & 0.400 & 7 & 0.170 & 0.228 & 0.095 & 0.833 \\
DINOv3 \citep{dinov3} & Whole Image & 0.400 & 5 & 0.168 & 0.225 & 0.094 & 0.813 \\
\midrule
EfficientNetV2 \citep{efficientnetv2} & Real-ESRGAN Upscaled & 0.500 & 89 & 0.146 & 0.580 & 0.128 & 0.170 \\
EfficientNetV2 \citep{efficientnetv2} & Whole Image & 0.500 & 88 & 0.145 & 0.609 & 0.224 & 0.107 \\
EfficientNetV2 \citep{efficientnetv2} & Lanczos Upscaled & 0.500 & 89 & 0.145 & 0.616 & 0.227 & 0.106 \\
EfficientNetV2 \citep{efficientnetv2} & Whole Image + WB + Bright & 0.500 & 95 & 0.142 & 0.634 & 0.209 & 0.108 \\
\midrule
ViT \citep{vit} & Whole Image + WB + Bright & 0.500 & 44 & 0.178 & 0.388 & 0.128 & 0.292 \\
ViT \citep{vit} & Real-ESRGAN Upscaled & 0.450 & 13 & 0.168 & 0.254 & 0.100 & 0.534 \\
ViT \citep{vit} & Whole Image & 0.450 & \underline{18} & 0.164 & 0.279 & 0.099 & 0.482 \\
ViT \citep{vit} & Lanczos Upscaled & 0.450 & 13 & 0.161 & 0.261 & 0.096 & 0.483 \\
\midrule
DeiT \citep{pmlr-v139-touvron21a} & Whole Image + WB + Bright & 0.500 & 38 & 0.160 & 0.428 & 0.125 & 0.223 \\
DeiT \citep{pmlr-v139-touvron21a} & Whole Image & 0.500 & 39 & 0.143 & 0.417 & 0.120 & 0.175 \\
DeiT \citep{pmlr-v139-touvron21a} & Lanczos Upscaled & 0.550 & 61 & 0.135 & 0.482 & 0.134 & 0.136 \\
DeiT \citep{pmlr-v139-touvron21a} & Real-ESRGAN Upscaled & 0.500 & 50 & 0.127 & 0.417 & 0.097 & 0.185 \\
\midrule
OSNet \citep{zhou2021osnet} & Whole Image + WB + Bright & 0.625 & 53 & 0.358 & 0.739 & 0.481 & 0.286 \\
OSNet \citep{zhou2021osnet}& Lanczos Upscaled & 0.625 & 61 & 0.358 & 0.779 & 0.556 & 0.264 \\
OSNet \citep{zhou2021osnet} & Whole Image & 0.625 & 63 & 0.358 & 0.783 & 0.552 & 0.265 \\
OSNet \citep{zhou2021osnet} & Real-ESRGAN Upscaled & 0.625 & 56 & 0.345 & 0.757 & 0.467 & 0.274 \\
\midrule
OSNet Fine-tuned & Whole Image & 0.643 & \textbf{19} & \textbf{0.831} & \textbf{0.895} & \textbf{0.806} & 0.858 \\
OSNet Fine-tuned & Real-ESRGAN Upscaled & 0.643 & 17 & \underline{0.815} & \textbf{0.895} & 0.775 & \underline{0.860} \\
OSNet Fine-tuned & Lanczos Upscaled & 0.643 & \underline{18} & 0.815 & 0.884 & \underline{0.786} & 0.845 \\
OSNet Fine-tuned & Whole Image + WB + Bright & 0.643 & \textbf{19} & 0.806 & \underline{0.888} & 0.770 & 0.847 \\
\bottomrule
\end{tabular}
\end{table}

\subsection{Object Re-Identification}
\label{app:object-reid-clustering}

We evaluate object re-identification using a labeled object dataset collected in the robot's office environment. The dataset composition is summarized in Table~\ref{tab:object_dataset_composition}. Since the detections contain some annotation noise, several identity labels appear under multiple detector classes. For example, the same physical object may be associated with different detector categories across samples.

The clustering results are reported in Table~\ref{tab:object_clustering_mean_std}. We compare different feature extraction models using three input variants: the original whole image, Lanczos upscaling~\citep{LanczosFilteringinOneandTwoDimensions}, and the original image with white balance and brightness adjustment. The results are averaged over multiple runs using greedy clustering and are reported as mean~$\pm$~standard deviation.

\begin{table}[H]
\centering
\scriptsize
\setlength{\tabcolsep}{3pt}
\renewcommand{\arraystretch}{1.08}
\caption{Class-wise composition of the labeled object dataset. Some identity labels appear under multiple detector classes due to annotation noise. All person names were replaced with pseudonymous labels.}
\label{tab:object_dataset_composition}
\resizebox{\linewidth}{!}{
\begin{tabular}{lrrp{0.58\linewidth}}
\toprule
\textbf{Class} & \textbf{Unique IDs} & \textbf{Labelled boxes} & \textbf{Identities} \\
\midrule
Chair & 6 & 59 & Chair-Kitchen (25), Chair-Alex (14), Chair-Blake (13), Chair-Casey (3), Chair-Dana (3), 3D-Printer (1) \\
Couch & 2 & 20 & Couch (19), Chair-Blake (1) \\
TV & 3 & 7 & Desktop-Alex (5), Desktop-Evan (1), Desktop-Casey (1) \\
Sports Ball & 1 & 8 & Volleyball (8) \\
Bed & 1 & 6 & Couch (6) \\
Oven & 2 & 4 & Chair-Kitchen (3), Rubbish (1) \\
Refrigerator & 1 & 2 & Rubbish (2) \\
Bottle & 2 & 3 & Dish-Soap (2), Bottle-Cola (1) \\
Cat / Car / Mouse / Toilet & 1 each & 1 each & Couch, Backpack-Evan, Volleyball, Rubbish-Blake \\
\bottomrule
\end{tabular}
}
\vspace{0.5em}
\begin{minipage}{0.95\linewidth}
\footnotesize
\textit{Note:} Some identity labels appear under multiple detector classes, e.g., Couch is labelled under both Couch and Bed detections, and Chair-Kitchen under Chair and Oven detections. This reflects annotation noise in the ground truth.
\end{minipage}
\end{table}

\begin{table}[H]
\centering
\scriptsize
\setlength{\tabcolsep}{3pt}
\renewcommand{\arraystretch}{1.08}
\caption{Object re-identification clustering results grouped by model and input variant. 
Values report mean $\pm$ standard deviation over runs using greedy clustering. 
Bold values indicate the best result per metric and underlined values indicate the second-best result.}
\label{tab:object_clustering_mean_std}
\begin{tabular}{llcccc}
\toprule
Model & Version & F1 & Purity & Precision & Recall \\
\midrule
CLIP \citep{pmlr-v139-radford21a} & Whole Image 
& $0.284 \pm 0.067$ & $0.297 \pm 0.115$ & $0.171 \pm 0.063$ & $0.973 \pm 0.065$ \\
CLIP \citep{pmlr-v139-radford21a} & Whole Image + WB + Bright 
& $0.283 \pm 0.058$ & $0.294 \pm 0.106$ & $0.167 \pm 0.044$ & $\underline{0.990 \pm 0.024}$ \\
CLIP \citep{pmlr-v139-radford21a} & Lanczos Upscaled 
& $0.271 \pm 0.033$ & $0.280 \pm 0.084$ & $0.159 \pm 0.028$ & $0.975 \pm 0.071$ \\

\midrule
ConvNeXt-Tiny \citep{convnext} & Whole Image 
& $\mathbf{0.734 \pm 0.102}$ & $0.914 \pm 0.057$ & $0.878 \pm 0.083$ & $0.659 \pm 0.151$ \\
ConvNeXt-Tiny \citep{convnext} & Lanczos Upscaled 
& $\underline{0.731 \pm 0.102}$ & $0.899 \pm 0.073$ & $0.868 \pm 0.101$ & $0.666 \pm 0.160$ \\
ConvNeXt-Tiny \citep{convnext} & Whole Image + WB + Bright 
& $0.720 \pm 0.127$ & $0.929 \pm 0.048$ & $0.901 \pm 0.071$ & $0.628 \pm 0.164$ \\

\midrule
DeiT \citep{pmlr-v139-touvron21a} & Whole Image 
& $0.551 \pm 0.222$ & $0.929 \pm 0.045$ & $0.867 \pm 0.074$ & $0.460 \pm 0.242$ \\
DeiT \citep{pmlr-v139-touvron21a} & Lanczos Upscaled 
& $0.555 \pm 0.224$ & $0.929 \pm 0.046$ & $0.870 \pm 0.075$ & $0.463 \pm 0.242$ \\
DeiT \citep{pmlr-v139-touvron21a} & Whole Image + WB + Bright 
& $0.483 \pm 0.173$ & $0.946 \pm 0.042$ & $0.882 \pm 0.080$ & $0.365 \pm 0.172$ \\

\midrule
DINOv3 \citep{dinov3} & Whole Image Embedding 
& $0.723 \pm 0.202$ & $0.878 \pm 0.075$ & $0.816 \pm 0.107$ & $0.744 \pm 0.280$ \\
DINOv3 \citep{dinov3} & Lanczos Upscaled 
& $0.722 \pm 0.202$ & $0.868 \pm 0.080$ & $0.804 \pm 0.107$ & $0.755 \pm 0.288$ \\
DINOv3 \citep{dinov3} & Whole Image + WB + Bright 
& $0.720 \pm 0.198$ & $0.885 \pm 0.076$ & $0.824 \pm 0.104$ & $0.735 \pm 0.290$ \\

\midrule
EfficientNetV2 \citep{efficientnetv2} & Whole Image 
& $0.458 \pm 0.193$ & $0.938 \pm 0.049$ & $0.857 \pm 0.086$ & $0.354 \pm 0.203$ \\
EfficientNetV2 \citep{efficientnetv2} & Lanczos Upscaled 
& $0.456 \pm 0.197$ & $0.941 \pm 0.044$ & $0.871 \pm 0.079$ & $0.347 \pm 0.201$ \\
EfficientNetV2 \citep{efficientnetv2} & Whole Image + WB + Bright 
& $0.465 \pm 0.152$ & $\mathbf{0.960 \pm 0.037}$ & $\mathbf{0.914 \pm 0.061}$ & $0.333 \pm 0.146$ \\

\midrule
OSNet \citep{zhou2021osnet} & Whole Image 
& $0.565 \pm 0.177$ & $0.794 \pm 0.156$ & $0.637 \pm 0.259$ & $0.723 \pm 0.301$ \\
OSNet \citep{zhou2021osnet}& Lanczos Upscaled 
& $0.557 \pm 0.177$ & $0.793 \pm 0.160$ & $0.637 \pm 0.266$ & $0.712 \pm 0.299$ \\
OSNet \citep{zhou2021osnet}& Whole Image + WB + Bright 
& $0.404 \pm 0.149$ & $0.649 \pm 0.296$ & $0.562 \pm 0.364$ & $0.631 \pm 0.323$ \\

\midrule
SAM2 \citep{ravi2025sam} & Whole Image 
& $0.361 \pm 0.000$ & $0.469 \pm 0.000$ & $0.222 \pm 0.000$ & $0.973 \pm 0.000$ \\
SAM2 \citep{ravi2025sam} & Lanczos Upscaled 
& $0.449 \pm 0.000$ & $0.478 \pm 0.000$ & $0.289 \pm 0.000$ & $\mathbf{0.998 \pm 0.000}$ \\
SAM2 \citep{ravi2025sam} & Whole Image + WB + Bright 
& $0.361 \pm 0.000$ & $0.469 \pm 0.000$ & $0.222 \pm 0.000$ & $0.973 \pm 0.000$ \\

\midrule
Swin \citep{liu2021Swin} & Whole Image 
& $0.655 \pm 0.212$ & $0.926 \pm 0.057$ & $0.905 \pm 0.064$ & $0.564 \pm 0.225$ \\
Swin \citep{liu2021Swin} & Lanczos Upscaled 
& $0.655 \pm 0.213$ & $0.926 \pm 0.056$ & $\underline{0.907 \pm 0.067}$ & $0.564 \pm 0.226$ \\
Swin \citep{liu2021Swin} & Whole Image + WB + Bright 
& $0.591 \pm 0.227$ & $0.938 \pm 0.055$ & $0.903 \pm 0.078$ & $0.493 \pm 0.232$ \\

\midrule
ViT \citep{vit} & Whole Image 
& $0.581 \pm 0.224$ & $0.828 \pm 0.117$ & $0.658 \pm 0.176$ & $0.633 \pm 0.269$ \\
ViT \citep{vit} & Lanczos Upscaled 
& $0.563 \pm 0.200$ & $0.816 \pm 0.106$ & $0.629 \pm 0.161$ & $0.629 \pm 0.263$ \\
ViT \citep{vit} & Whole Image + WB + Bright 
& $0.513 \pm 0.228$ & $0.900 \pm 0.049$ & $0.737 \pm 0.083$ & $0.459 \pm 0.266$ \\
\bottomrule
\end{tabular}
\end{table}

\FloatBarrier
\newpage
\section{Fine-Tuning}
\label{app:finetune}

We fine-tune OSNet from an MSMT17-pretrained \citep{8578114} checkpoint using manually labeled person detections collected from the Spot robot's front fisheye camera. 
The training dataset contains 808 labeled detections from 19 identities. 
We split the dataset into 605 training samples and 203 validation samples. 
Training uses identity-balanced batches, AdamW \citep{loshchilov2018decoupled}, a cosine learning-rate schedule, and a combined cross-entropy and triplet loss. 
We select the best checkpoint at epoch 24 using validation pairwise F1, with validation loss as a tie-breaker.

\begin{table}[H]
  \centering
  \scriptsize
  \setlength{\tabcolsep}{6pt}
  \renewcommand{\arraystretch}{1.1}
  \caption{Training setup and hyperparameters used for fine-tuning OSNet on the person re-identification dataset.}
  \label{tab:osnet_finetuning_hyperparameters}
  \begin{tabular}{ll}
  \toprule
  \textbf{Parameter} & \textbf{Value} \\
  \midrule
  Base checkpoint & MSMT17-pretrained OSNet x1.0 \\
  Camera source & Spot front fisheye camera \\
  Input size & $128 \times 256$ (W$\times$H) \\
  Feature dimension & $512$ \\
  Dataset size & $808$ labeled detections \\
  Identities & $19$ \\
  Training samples & $605$ \\
  Validation samples & $203$ \\
  Best checkpoint & Epoch $24$ \\
  Checkpoint selection & validation pairwise F1; validation loss as tie-breaker \\
  Optimizer & AdamW  \\
  Learning-rate schedule & Cosine \\
  Loss & Cross-Entropy (label smoothing $0.1$) + triplet loss \\
  Batch sampling & Identity-balanced \\
  \texttt{batch\_size} & $32$ \\
  \texttt{bbox\_margin} & $0.08$ \\
  \texttt{epochs} & $40$ \\
  \texttt{freeze\_epochs} & $40$ \quad (early layers frozen for all epochs) \\
  \texttt{instances\_per\_id} & $4$ \\
  \texttt{lr} & $0.0003$ \\
  \texttt{min\_samples\_per\_id} & $2$ \\
  \texttt{num\_workers} & $2$ \\
  \texttt{seed} & $42$ \\
  \texttt{triplet\_margin} & $0.3$ \\
  \texttt{triplet\_weight} & $1.2$ \\
  \texttt{val\_frac} & $0.25$ \\
  \texttt{weight\_decay} & $0.0005$ \\
  \bottomrule
  \end{tabular}
  \end{table}

\end{document}